\documentclass{article}

\usepackage{microtype}
\usepackage{graphicx}
\usepackage{booktabs} %

\usepackage{hyperref}

\usepackage{siunitx}

\usepackage[accepted]{icml2024}

\usepackage{amsmath}
\usepackage{amssymb}
\usepackage{mathtools}
\usepackage{amsthm}

\theoremstyle{plain}
\newtheorem{theorem}{Theorem}[section]

\theoremstyle{definition}

\theoremstyle{remark}
\newtheorem{remark}[theorem]{Remark}

\newif\ifshowtmp
\showtmptrue
\showtmpfalse

\newif\ifarxiv
\arxivtrue

\newif\ifshowtmp
\showtmptrue
\showtmpfalse

\ifshowtmp
    \newcommand{\todom}[1]{{\textcolor{usccardinal}{(todo minor: #1)}}}
	\newcommand{\todo}[1]{{\textcolor{red}{(TODO: #1)}}}
    \newcommand{\tocheck}[1]{{\textcolor{red}{(TO CHECK: #1)}}}

	\newcommand{\T}[1]{\textcolor{usccardinal}{#1}}

    \makeatletter
    \def\w#1 {\T{#1}~}
    \makeatother

\else
    \newcommand{\todom}[1]{}
    \newcommand{\tocheck}[1]{}
	
	\newcommand{\T}{}
	\newcommand{\todo}[1]{}

    \newcommand{\w}{}
\fi
\newcommand{\todoLessImportant}[1]{}
\newcommand{\removeForSavingSpace}[1]{}

\newif\ifPPT
\PPTtrue

\newif\ifshowstruct
\showstructtrue
\showstructfalse %

\ifshowstruct
	\newcommand{\tstructz}[1]{[#1] }
\else
	\newcommand{\tstructz}[1]{}
\fi

\label{packages}
\usepackage{mfirstuc}
\usepackage{graphicx}%
\usepackage{subcaption}
\usepackage{caption}

\usepackage{multirow}
\usepackage{multicol}
\usepackage{array}
\usepackage{booktabs}%

\usepackage{amsmath}
\usepackage{amssymb, graphicx, url, mathtools}%
\usepackage[overload]{empheq}

\usepackage{algorithmic,algorithm}   %

\usepackage{footnote}
\makesavenoteenv{tabular}
\makesavenoteenv{table}
\usepackage{pdfpages}
\usepackage{textcomp}
\usepackage{chngcntr}
\usepackage{verbatim}

\usepackage{xparse,calc}
\usepackage{afterpage}
\usepackage{comment}

\usepackage{hyperref}
\usepackage[capitalize,noabbrev,nameinlink]{cleveref}

\hypersetup{
    colorlinks,
    linkcolor={black},
    citecolor={black},
    urlcolor={black}
}

\label{Command}

\counterwithin*{question}{section}

\makeatletter
\def\renewtheorem#1{%
  \expandafter\let\csname#1\endcsname\relax
  \expandafter\let\csname c@#1\endcsname\relax
  \gdef\renewtheorem@envname{#1}
  \renewtheorem@secpar
}
\def\renewtheorem@secpar{\@ifnextchar[{\renewtheorem@numberedlike}{\renewtheorem@nonumberedlike}}
\def\renewtheorem@numberedlike[#1]#2{\newtheorem{\renewtheorem@envname}[#1]{#2}}
\def\renewtheorem@nonumberedlike#1{  
\def\renewtheorem@caption{#1}
\edef\renewtheorem@nowithin{\noexpand\newtheorem{\renewtheorem@envname}{\renewtheorem@caption}}
\renewtheorem@thirdpar
}
\def\renewtheorem@thirdpar{\@ifnextchar[{\renewtheorem@within}{\renewtheorem@nowithin}}
\def\renewtheorem@within[#1]{\renewtheorem@nowithin[#1]}
\makeatother

\renewtheorem{theorem}{Theorem}
\renewtheorem{remark}{Remark}
\renewtheorem{definition}{Definition}
\renewtheorem{example}{Example}

\crefname{mdpexample}{MDP Example}{MDP Examples}

\crefname{todo}{TODO}{TODO}

\makeatletter
\newcommand{\dummylabel}[2]{\def\@currentlabel{#2}\label[td]{#1}}
\makeatother

\usepackage{thmtools} 
\usepackage{thm-restate}

\usepackage{xstring}

\newcolumntype{+}{>{\global\let\currentrowstyle\relax}}
\newcolumntype{Y}{>{\currentrowstyle}}\label{key}
\newcolumntype{-}{>{\currentrowstyle}}

\newcommand{\breakingcomma}{
  \begingroup\lccode`~=`,
  \lowercase{\endgroup\expandafter\def\expandafter~\expandafter{~\penalty0 }}
}

\DeclareFontFamily{U}{mathx}{\hyphenchar\font45}
\DeclareFontShape{U}{mathx}{m}{n}{<-> mathx10}{}
\DeclareSymbolFont{mathx}{U}{mathx}{m}{n}
\DeclareMathAccent{\widebar}{0}{mathx}{"73}

\renewcommand{\algorithmicrequire}{\textbf{Input:}}
\renewcommand{\algorithmicensure}{\textbf{Output:}}

\label{Settings}

\usepackage{makecell}

\Crefname{figure}{Fig.}{Fig.}
\Crefname{equation}{Eq.}{Eq.}
\Crefname{assumption}{Assumption}{assumption}

\newcommand{\crefnop}[1]{\namecref{#1} \ref{#1}}

\newcommand{\Crefnop}[1]{\nameCref{#1} \ref{#1}}

\newcommand{\reftodo}[1]{\todo{cref}}

\newcommand{\Creftodo}[1]{\todo{cref}}
\newcommand{\creftodo}[1]{\todo{cref}}
\newcommand{\citetodo}[1]{\todo{cite}}

\newcommand{\Creftd}[1]{\todo{cref}}
\newcommand{\creftd}[1]{\todo{cref}}
\newcommand{\citetd}[1]{\todo{cite}}

\usepackage[export]{adjustbox}

\usepackage{xcolor}
\definecolor{bluemy}{HTML}{1f77b4}

\usepackage{enumitem}

    \usepackage{xr}
    \makeatletter
    \newcommand*{\addFileDependency}[1]{
      \typeout{(#1)}
      \@addtofilelist{#1}
      \IfFileExists{#1}{}{\typeout{No file #1.}}
    }
    \makeatother

\makeatletter
\def\widebreve{\mathpalette\wide@breve}
\def\wide@breve#1#2{\sbox\z@{$#1#2$}%
     \mathop{\vbox{\m@th\ialign{##\crcr
\kern0.08em\brevefill#1{0.8\wd\z@}\crcr\noalign{\nointerlineskip}%
                    $\hss#1#2\hss$\crcr}}}\limits}
\def\brevefill#1#2{$\m@th\sbox\tw@{$#1($}%
  \hss\resizebox{#2}{\wd\tw@}{\rotatebox[origin=c]{90}{\upshape(}}\hss$}
\makeatletter

\usepackage{accents}

\usepackage[makeroom]{cancel}

\definecolor{blueMy}{HTML}{1f77b4}
\definecolor{orangeMy}{HTML}{ff7f0e}
\definecolor{greenMy}{HTML}{2ca02c}
\definecolor{redMy}{HTML}{d62728}
\definecolor{purpleMy}{HTML}{9467bd}
\definecolor{brownMy}{HTML}{8c564b}
\definecolor{usccardinal}{rgb}{0.6, 0.0, 0.0}

\usepackage{makecell}

\NewDocumentCommand{\pooling}{ O{} O{} }{ \mathbb{P} }

\NewDocumentCommand{\conv}{ O{} O{} }{ \mathbb{C} }

\NewDocumentCommand{\stack}{ O{} O{} }{ {#1} \oplus {#2} }

\NewDocumentCommand{\linear}{ O{} O{} }{\mathbb{L} }

\def\policyIndex{ n_{p} }
\def\policyCount{ N_{p} }

\def\blockIndex{ n_{B} }
\def\blockCount{ N_{B} }

\def\blockDepth{ N_{b} }

\NewDocumentCommand{\policy}{ O{ ( n +n_b )} O{\policyIndex} }{ {\overline{\pi}}^{#1}_{#2} }

\def\latentASpace{ \overline{\mathcal{A}} }

\def\latentA{ \overline{a} }

\NewDocumentCommand{\V}{ O{} O{} }{ \overline V^{(#1)}_{#2}  }
\NewDocumentCommand{\Q}{ O{} O{} }{ \overline Q^{(#1)}_{#2}  }
\NewDocumentCommand{\PiM}{ O{} O{} }{ \overline \Pi^{(#1)}_{#2}  }

\newcommand{\vfVE}{  \mathcal }

\NewDocumentCommand{\characteristic}{ O{} }{\emph{(#1)}}

\def\softmax{\mathop{smax}\nolimits^{\alpha}}

\NewDocumentCommand{\nstepReturn}{O{Q} O{n}   }{ {G}_{#1}^{#2} }
\NewDocumentCommand{\nstepReturnAvg}{O{Q} O{n}   }{ {\bar G}_{#1}^{#2} }

\def\lanpaper/{[Lan, 2020]}
\def\youngpaper/{[Young, 2019]}

\NewDocumentCommand{\K}{ O{k} }{ _{#1} }

\definecolor{expectationPolicyStep}{named}{black}
\definecolor{maxStepOneN}{named}{red}
\definecolor{maxStep}{named}{maxStepOneN}
\definecolor{maxPolicyStep}{named}{blue}
\definecolor{softmaxPolicyStep}{named}{black}
\definecolor{stepFirst}{named}{pink}
\NewDocumentCommand{\EPolicyStep}{ O{expectationPolicyStep}  }
{{
    \color{#1}
    \E_{ \pi \sim \policyDist,  n  \sim \stepDist }
}}

\NewDocumentCommand{\SoftmaxPolicySoftmaxStep}{ }
{{
    { \color{softmaxPolicyStep} 
    \underset{\pi\in \policySet } {\mathop {smax}\nolimits^{ \widetilde \alpha} }
    }
   { 
   \color{softmaxPolicyStep}
    \underset{n\in  \stepSet } {\mathop {smax}\nolimits^{\alpha} }
   }
}}

\NewDocumentCommand{\SoftmaxPolicySoftmaxStepNoSet}{ }
{{
    { \color{softmaxPolicyStep} 
    \underset{\pi } {\mathop {smax}\nolimits^{ \widetilde \alpha} }
    }
   { 
   \color{softmaxPolicyStep}
    \underset{n } {\mathop {smax}\nolimits^{\alpha} }
   }
}}

\NewDocumentCommand{\dataset}{ O{m} O{s,a} }{ {\cal D}_{#2}^{\ifthenelse{\equal{#1}{}}{}{(#1)}} }
\NewDocumentCommand{\traj}{ O{\pi} O{s_t,a_t} }{ {\tau}_{#2}^{#1} }
\NewDocumentCommand{\trajnew}{ O{\pi} O{s_t,a_t} O{n} }{ {\tau}_{#2}^{#3} \sim {#1} }

\def\step/{lookahead depth}
\def\steps/{lookahead depths}
\def\stepText/{lookahead depth}
\def\stepsText/{lookahead depths}

\def\stepSet{{\cal N}}
\def\stepSetMath/{ $\stepSet$ } 

\def\stepSetText/{set of \steps/}
\def\stepSetTextMath/{set of \steps/ $\stepSet$}
\def\StepSetText/{Set of \steps/}

\def\policyText/{lookahead policy}
\def\policiesText/{lookahead policies}
\def\policySetText/{set of lookahead policies}
\def\PolicySetText/{Set of lookahead policies}
\def\policyDistText/{selection distribution of behavioral policies }

\NewDocumentCommand{\policyDist}{ O{} O{\policySet} }{ {P}^{#1}_{#2} }

\NewDocumentCommand{\stepDist}{ O{} O{\stepSet} }{ {P}^{#1}_{#2} }

\NewDocumentCommand{\policySet}{ }{ {\Pi} }

\def\policySetMath/ { $\policySet$ }

\def\policySetTextMath/{\policySetText/ $\policySet$}
\def\policyDistTextMath/{\policyDistText/ $\policyDist$}

\NewDocumentCommand{\dist}{ O{} O{} }{ \mathcal{P}^{#1}_{#2} }

\NewDocumentCommand{\bellmanOperator}{ O{} O{} }{ \mathcal{B}^{#1}_{#2} }

\NewDocumentCommand{\bellmanOptOperator}{ O{} O{} }{ \mathcal{B}^{#1}_{#2} }
\NewDocumentCommand{\BOOperator}{ O{} O{} }{ \mathcal{B}^{#1}_{#2} }
\NewDocumentCommand{\BOOperatorEmbed}{ O{} O{} }{ \mathcal{B}^{#1}_{#2} }
\NewDocumentCommand{\BEOperator}{ O{\pi} O{} }{ \mathcal{B}^{#1}_{#2} }
\NewDocumentCommand{\BEOperatorEmbed}{ O{\overline{\pi}} O{} }{ \mathcal{B}^{#1}_{#2} }
\NewDocumentCommand{\multistepBOOperator}{ O{\policyDist} O{\stepDist} }{\overline{\mathcal{B}}^{#1}_{#2} }
\def\multistepBOOperatorText/{Multi-step BO Operator}
\def\multistepBOOperatorTextMath/{Multi-step BO Operator $\multistepBOOperator[][]$}

\def\multistepBEOperatorText/{Multi-step IS-based BE Operator}
\NewDocumentCommand{\multistepBEOperator}{ O{\policyDist} O{\stepDist} O{} }{ \overset{ {\tiny #3} }{ \widebreve{ \mathcal{B} } }^{#1}_{#2} }
\def\multistepBEOperatorTextMath/{\multistepBEOperatorText/ $\multistepBEOperator[][]$}

\def\stStepSet/{\text{s.t. }}

\def\highwayPrefix/{Highway}
\def\highwayPrefixFull/{Highway}

\def\highwayQLearning/{\highwayPrefix/ Q-Learning}
\def\highwayDQN/{\highwayPrefix/ DQN}

\def\softHighwayDQN/{Soft \highwayPrefix/ DQN}
\def\nstepHighwayDQN/{$n$-step \highwayPrefix/ DQN}

\def\highwayQLearningFull/{\highwayPrefixFull/ Q-Learning}
\def\highwayDQNFull/{\highwayPrefixFull/ DQN}

\def\highwayValueIteration/{\highwayPrefixFull/ Value Iteration}

\def\highwayOperatorText/{\highwayPrefix/ Operator}

\def\highwayGenOperatorText/{\highwayPrefix/ Generalized Operator}

\NewDocumentCommand{\highwayOperator}{ O{\policyDist} O{\stepDist} }{   {{{\mathcal{G}}}}^{ {#1} }_{ {#2} }   }

\NewDocumentCommand{\highwayOperatorQ}{ O{\policyDist} O{\stepDist} }{   {{\mathcal{G}}}^{ {#1} }_{ {#2} }   }
\NewDocumentCommand{\wronghighwayOperator}{ O{\policyDist} O{\stepDist} }{   {\bcancel{{\mathcal{G}}} }^{ {#1} }_{ {#2} }   }

\NewDocumentCommand{\highwayOptOperator}{ O{\policySet} O{\stepSet} }{   {\accentset{\mbox{\large\bfseries .}}{\mathcal{G}}}^{ {#1} }_{ {#2} }   }
\def\highwayOptOperatorText/{Highway Optimality Operator}
\def\highwayOptOperatorTextMath/{Highway Optimality Operator $\highwayOptOperator[][]$}
\def\highwayOperatorTextMath/{\highwayOperatorText/ $\highwayOperator[][]$}
\def\highwayGenOperatorTextMath/{\highwayGenOperatorText/ $\highwayOperator[][]$}

\def\highwaySoftmaxOperatorText/{\highwayPrefix/ Softmax Operator}
\NewDocumentCommand{\highwaySoftmaxOperator}{ O{\stepSet, \alpha} O{\policySet,\widetilde\alpha} }{   {{\mathcal{G}}}_{ {#1} }^{ {#2} }   }
\def\highwaySoftmaxOperatorTextMath/{\highwaySoftmaxOperatorText/ $\highwaySoftmaxOperatorMath$}

\def\highwayEquation/{\highwayPrefix/ Equation}
\def\highwayEquationFull/{\highwayPrefixFull/ Bellman Optimality Equation}

\def\QQQQ{}

\def\highwayQQPrefix/{\highwayPrefix/ $Q$}
\def\highwayQQPrefixFull/{\highwayPrefixFull/ $Q$}
\def\highwayQQOperatorText/{\highwayQQPrefix/ Operator}
\def\highwayQQOperatorTextFull/{\highwayQQPrefixFull/ Bellman Optimality Operator}
\def\highwayQQOperatorTextMath/{\highwayQQOperatorText/ $\highwayQQOperator$}

\def\highwayQQEquation/{\highwayQQPrefix/ Equation}
\def\highwayQQEquationFull/{\highwayQQPrefixFull/ Bellman Optimality Equation}

\def\greedyReturn/{\highwayPrefix/ return}
\def\GreedyReturn/{\HighwayPrefix Return}

\NewDocumentCommand{\distance}{ O{} }{ \mathbf{d}_{#1}^Q(s,a) } 
\NewDocumentCommand{\distanceAll}{ O{} }{ d_{#1}^{Q} } 

\NewDocumentCommand{\assumptionAn}{ O{n} O{\widehat\Pi} O{\pi^*} }{Assumption A$(#2,#1)$}

\NewDocumentCommand{\stateSetNStep}{ O{s} O{\pi} O{n} }{ \mathcal{S}_{#1, #3}^{#2} }

\NewDocumentCommand{\ourExpOpt}{ O{\widehat{\Pi}} O{\mathcal{N}} }{ \overline{\mathcal{G}}_{#1}^{#2} }
\def\ourExpOptText/{Expectation Highway Operator}

\def\oneStepOperatorText/{BO operator}
\NewDocumentCommand{\oneStepOperatorMath}{ O{}  }{ \multiStepOperatorMath[#1][] }
\NewDocumentCommand{\oneStepOperator}{ O{}  }{ \multiStepOperatorMath[#1][] }
\def\oneStepOperatorTextMath/{\oneStepOperatorText/ $\oneStepOperatorMath$}

\def\bellmanOptText/{Bellman Optimality Operator}
\NewDocumentCommand{\bellmanOpt}{ O{}  }{ \mathcal{B}^{#1} }
\def\bellmanExpText/{Bellman Expectation Operator}
\NewDocumentCommand{\bellmanExp}{ O{#1}  }{ \bellmanOpt[#1] }
\NewDocumentCommand{\bellmanIS}{ O{} O{}  }{{\breve{\mathcal{B}}}^{#1}_{#2} }

\def\oneStepQQOperatorText/{Bellman Optimality Operator}
\NewDocumentCommand{\oneStepQQOperator}{ O{}  }{ \multiStepQQOperator[#1][] }
\NewDocumentCommand{\oneStepQQOperatorMath}{ O{}  }{ \multiStepQQOperator[#1][] }
\def\oneStepQQOperatorTextMath/{\oneStepQQOperatorText/ $\oneStepQQOperatorMath$}

\NewDocumentCommand{\multiStepOnPolicyOperator}{ O{N} O{\pi} }{ {\mathcal{B}}^{{#2}}_{#1} }
\NewDocumentCommand{\multiStepOnPolicyOperatorMath}{ O{N} O{\pi} }{ \multiStepOnPolicyOperator[#1][#2] }

\def\multiStepOnPolicyOperatorText/{Multi-Step Bellman Expectation Operator}
\def\multiStepOnPolicyOperatorTextMath/{ \multiStepOnPolicyOperatorText/ $\multiStepOnPolicyOperatorMath$ }

\NewDocumentCommand{\multiStepOnPolicyQQOperatorMath}{ O{N} O{\pi} }{ \QQQQ{\mathcal{B}}^{{#2}}_{#1} }

\NewDocumentCommand{\multiStepOffPolicyOperatorMath}{ O{N} O{\pi} O{\policyDist} }{ {\mathcal{B}}^{{#2}}_{#1} }
\NewDocumentCommand{\multiStepOffPolicyQQOperatorMath}{ O{N} O{\pi} O{\policySet} }{ \QQQQ{\mathcal{B}}^{{#2,#3}}_{#1} }

\NewDocumentCommand{\nreturnWeight}{ O{n} O{\pi}  }{ w_{#1}^{#2} }

\def\multiStepOperatorText/{Multi-step Bellman Optimality operator}
\NewDocumentCommand{\multiStepOperator}{ O{N} O{\policySet} }{ \mathcal{B}^{{#1}}_{#2} }
\NewDocumentCommand{\multiStepOperatorMath}{ O{N} O{\policySet} }{ \multiStepOperator[#1][#2] }
\def\multiStepOperatorTextMath/{\multiStepOperatorText/ $\multiStepOperatorMath$}

\def\multiStepQQOperatorText/{\multiStepOperatorText/}
\NewDocumentCommand{\multiStepQQOperator}{ O{N} O{\policyDist} }{ \QQQQ{\mathcal{B}}^{{#1}}_{#2} }
\NewDocumentCommand{\multiStepQQOperatorMath}{ O{N} O{\pi} }{ \multiStepQQOperator[#2][#1] }
\def\multiStepQQOperatorTextMath/{\multiStepOperatorText/ $\multiStepQQOperatorMath$}

\def\highwayPolicyIteration/{\highwayPrefixFull/ Policy Iteration}

\def\HighwayEvaluatePrefix/{Highway}
\def\highwayEvaluatePrefix/{Highway}
\def\highwayEvaluatePrefixFull/{Highway}
\def\highwayEvaluateValueIteration/{\highwayPrefixFull/ Value Iteration}
\def\highwayEvaluateEquationFull/{\highwayPrefixFull/ Recombined Equation}

\def\highwayEvaluateEquation/{\highwayPrefix/ Recombined Equation}
\def\highwayEvaluateOperatorText/{\highwayPrefix/ Recombined Operator}
\def\highwayEvaluateOperatorTextFull/{\highwayPrefixFull/ Recombined Operator}
\NewDocumentCommand{\highwayEvaluateOperatorMath}{ O{N} O{\policySet} }{ \mathcal{G}^{ {#1} }_{ {#2} } }

\def\recombinedPolicySetText/{set of recombined behavioral policies}

\NewDocumentCommand{\highwayRecombinationOperatorMath}{ O{N} }{ \mathcal{G}^{{#1}} }

\definecolor{QPiFixedPoint}{named}{usccardinal}

\def\nrandomseed/{5}

\usepackage{amsmath,amsfonts,bm}

\def\eqref#1{equation~\ref{#1}}

\def\1{\bm{1}}

\DeclareMathAlphabet{\mathsfit}{\encodingdefault}{\sfdefault}{m}{sl}
\SetMathAlphabet{\mathsfit}{bold}{\encodingdefault}{\sfdefault}{bx}{n}

\newcommand{\E}{\mathbb{E}}

\newcommand{\R}{\mathbb{R}}

\DeclareMathOperator*{\argmax}{arg\,max}

\NewDocumentCommand{\rangeInt}{ O{a} O{b} }{ [ #1..#2 ] }

\ifshowtmp
    \newcommand{\yuhui}[1]{\textcolor{teal}{(\textbf{Yuhui}: #1})}
    \newcommand{\francesco}[1]{\textcolor{orange}{(\textbf{Francesco:} #1})}
    \newcommand{\weida}[1]{\textcolor{blue}{(\textbf{Weida}: #1)}}
    \newcommand{\qingyuan}[1]{\textcolor{purple}{(\textbf{Qingyuan}: #1)}}
    \newcommand{\sarah}[1]{\textcolor{blue}{(\textbf{Sarah}: #1)}}
\else
    \newcommand{\yuhui}[1]{}
    \newcommand{\francesco}[1]{}
    \newcommand{\weida}[1]{}
    \newcommand{\qingyuan}[1]{}
    \newcommand{\sarah}[1]{}
\fi

\icmltitlerunning{Highway Value Iteration Networks}

\begin{document}

\twocolumn[
\icmltitle{Highway Value Iteration Networks}

\icmlsetsymbol{equal}{*}

\begin{icmlauthorlist}
\icmlauthor{Yuhui Wang}{equal,KAUST}
\icmlauthor{Weida Li}{equal,NUS}
\icmlauthor{Francesco Faccio}{KAUST,IDSIA}
\icmlauthor{Qingyuan Wu}{Liverpool}
\icmlauthor{Jürgen Schmidhuber}{KAUST,IDSIA}
\end{icmlauthorlist}

\icmlaffiliation{KAUST}{AI Initiative, King Abdullah University of Science and Technology}
\icmlaffiliation{NUS}{National University of Singapore}
\icmlaffiliation{IDSIA}{The Swiss AI Lab IDSIA/USI/SUPSI}
\icmlaffiliation{Liverpool}{The University of Liverpool}

\icmlcorrespondingauthor{Yuhui Wang}{yuhui.wang@kaust.edu.sa}

\icmlkeywords{Machine Learning, ICML}

\vskip 0.3in
]

\printAffiliationsAndNotice{\icmlEqualContribution} %

\begin{abstract}

Value iteration networks (VINs) enable end-to-end learning for planning tasks by employing a differentiable ``planning module'' that approximates the value iteration algorithm. However, long-term planning remains a challenge because training very deep VINs is difficult. To address this problem, we embed highway value iteration---a recent algorithm designed to facilitate long-term credit assignment---into the structure of VINs. This improvement augments the ``planning module" of the VIN with three additional components: 
1) an ``aggregate gate,'' which constructs skip connections to improve information flow across many layers;
2) an ``exploration module," crafted to increase the diversity of information and gradient flow in spatial dimensions;
3) a ``filter gate" designed to ensure safe exploration.
The resulting novel \emph{highway VIN} can be trained effectively with hundreds of layers using standard backpropagation. 
In long-term planning tasks requiring hundreds of planning steps, deep highway VINs outperform both traditional VINs and several advanced, very deep NNs.

\end{abstract}

\section{Introduction}\label[todo]{todo}

Planning is a search for action sequences that are predicted to achieve specific goals. 
The value iteration network (VIN) \cite{tamar2016value} is a neural network (NN) architecture that enables end-to-end training for planning tasks using an embedded ``planning module,'' a differentiable approximation of the value-iteration algorithm \cite{bellman1966dynamic}.
VINs have exhibited remarkable proficiency in various tasks, including 
path planning \cite{pflueger2019rover,jin2021value}, autonomous navigation \cite{wohlke2021hierarchies}, and complex decision-making in dynamic environments \cite{li2021dynamic}.

\begin{figure}
    \centering
    \includegraphics[width=0.6\linewidth]{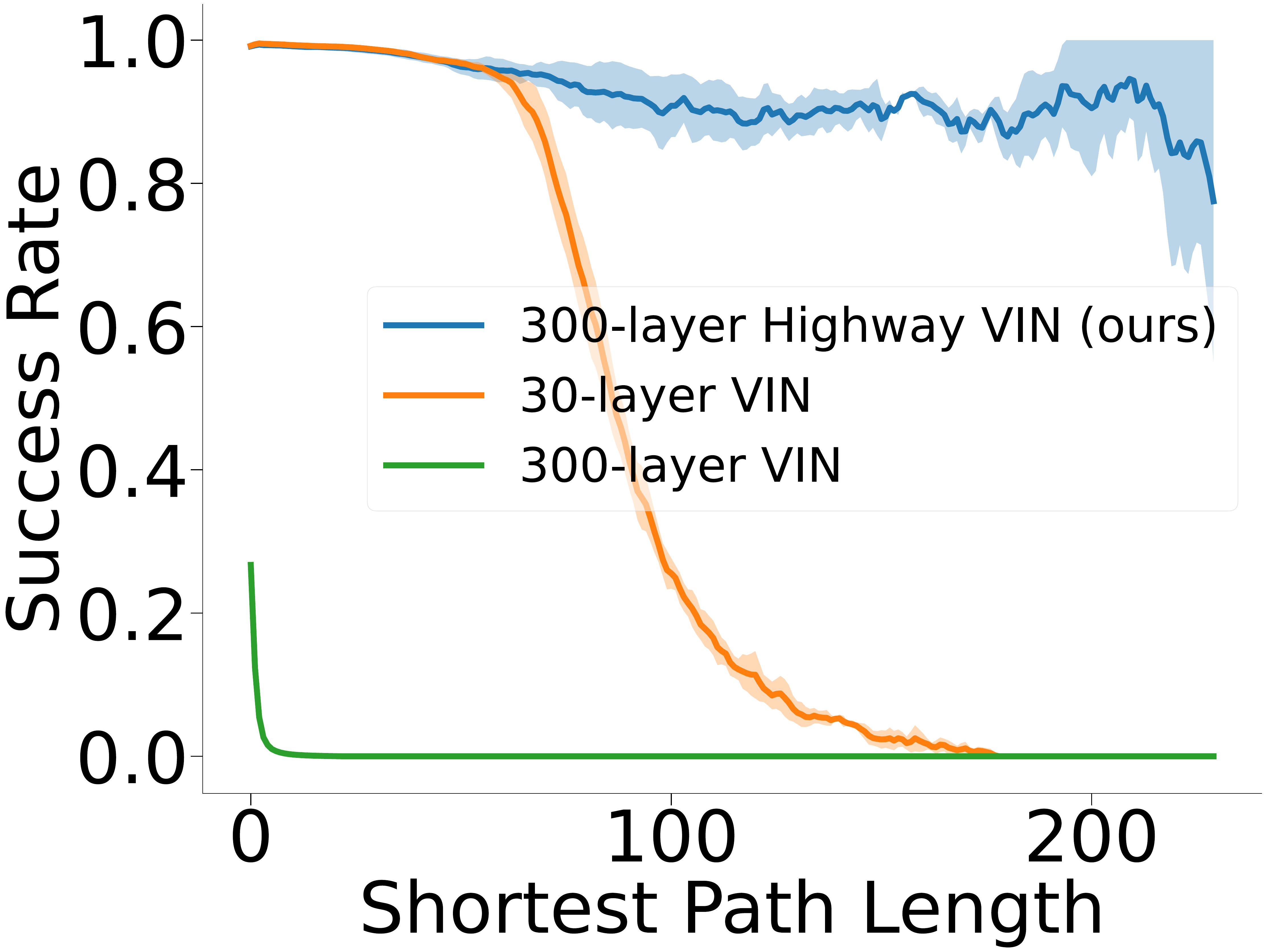}
    \caption{
    Success rates of reaching the goal in a $25\times 25$ maze problem.
    The success rate of a 30-layer VIN considerably decreases as the shortest path length increases, and training a 300-layer VIN is difficult and exhibits poor performance.
    }
    \label{fig_introduction}
\end{figure}

However, VIN encounters significant challenges in long-term planning. 
For example, in path planning tasks where the shortest path length exceeds 120, the success rate of VINs in reaching the goal drastically decreases below 10\% (\Cref{fig_introduction}).
A promising approach to improve the long-term planning capabilities of VIN is increasing the depth of its embedded ``planning module.''
A deeper planning module can integrate more planning steps in VINs, potentially improving their ability to perform long-term planning.

Nonetheless, increasing the depth of an NN can introduce complications, such as vanishing or exploding gradients, which is a fundamental problem in deep learning \cite{Hochreiter:91}.
Although very deep NNs can be effectively trained in classification tasks using different methods 
\cite{srivastava2015training, srivastava2015highway, he2016deep, huang2017densely}, these methods have not been equally successful in planning tasks (\Cref{table_success_rate} in \Cref{sec_experiment}). 
This disparity may arise from the unique architecture of the VIN, particularly its planning module, 
which incorporates a specific inductive bias from reinforcement learning (RL). This inductive bias is grounded in the value iteration algorithm, known for its theoretical soundness \cite{bellman1966dynamic, sutton2018reinforcement}. 
Herein, additional RL-relevant prior knowledge is integrated into the VIN architecture to address these challenges.

This study aims to integrate well-established techniques from both the fields of NNs and RL to create an effective and theoretically sound method.
Central to the foundation of this proposed approach is highway value iteration \cite{wang2023highway}. This method was designed to facilitate efficient long-term credit assignment in the context of RL.
We also leverage the architectural innovations of highway networks \cite{srivastava2015training, srivastava2015highway} and their variants residual networks \citep{he2016deep} and DenseNets \citep{huang2017densely}, particularly their use of skip connections, which are advantageous for training exceptionally deep NNs.

Building on this foundation, the VIN is enhanced by embedding the highway value iteration algorithm into its core ``planning module.'' 
This integration introduces three key innovations:
(1) an ``\emph{aggregate gate}'' for creating skip connections and improving information flow between layers;
(2) an ``\emph{exploration module}'' that injects controlled stochasticity during training, thereby diversifying information and gradient flow across spatial dimensions;
(3) and a "\emph{filter gate}" designed to ``filter out'' useless exploration paths, ensuring safe and efficient exploration.
These improvements result in the \emph{highway value iteration network (highway VIN)}, a new VIN variant specifically tailored for long-term planning. 
Remarkably, highway VINs can be efficiently trained with hundreds of layers using standard backpropagation techniques.
This study highlights the connections between highway RL and highway networks and their combined potential to advance the capabilities of deep learning models in complex planning tasks.
Notably, in scenarios requiring extensive planning, highway VINs outperform traditional VINs and several advanced deep NN models. This showcases their superior capability in handling complex, long-term planning challenges. 
The source code of highway VIN is available at \href{https://github.com/wangyuhuix/HighwayVIN}{https://github.com/wangyuhuix/HighwayVIN}.

\section{Related Work}

\paragraph{Variants of Value Iteration Networks.}
VINs \cite{tamar2016value} are important architectures that integrate planning capabilities directly into NNs.
VINs have a notable advantage over classical RL methods, as they learn policies that generalize better on novel tasks.
However, VINs are challenged by issues such as training instability, hyperparameter sensitivity, and overestimation bias.
These issues can be addressed using gated path-planning networks (GPPN) \cite{lee2018gated}, a recurrent version of the VIN, which replaces convolutional networks with gated recurrent networks, resulting in more stable and effective learning.
For higher-dimensional planning tasks, AVINs \cite{schleich2019value} extend VINs with multi-level abstraction modules. These modules can capture various types of useful information during learning.
Another significant challenge arises in generalizing VINs to target domains with limited data. Transfer VINs \cite{shen2020transfer}  tackle this issue by proposing a transfer learning approach, effectively adapting VINs to different, unseen target domains.
Unfortunately, existing methods are considerably limited to extend to real-world and large-scale planning problems as shallow NNs lack long-term planning ability.

\paragraph{Neural Networks with Deep Architectures.} 

Deep learning involves assigning credits to NN components that affect the performance of the NN across multiple layers, or in the case of sequential data, over several time steps. 
In 1965, \citet{ivakhnenko1965} introduced the first learning algorithms for deep feedforward NNs (FNNs) with any number of hidden layers.
However, training FNNs with more than six layers by gradient descent remained a challenge until the early 2010s \cite{ciresan:2010}. 
Similarly, in the 1980s, recurrent neural networks (RNNs) were limited to problems spanning fewer than ten time steps 
due to the ``vanishing gradient problem'' \cite{Hochreiter:91}, the fundamental problem of deep learning. In very deep NNs, and in RNNs processing sequences with significant time lags between relevant events, the backpropagated gradients tend to either explode or vanish.
In 1991, advances in history compression and neural sequence chunking through self-supervised pre-training enabled training RNNs over hundreds or thousands of steps \cite{Schmidhuber:91chunker, Schmidhuber:92ncchunker}. However, this worked only for sequences with predictable regularities. 
This limitation was overcome using residual recurrent connections \cite{Hochreiter:91} in long short-term memory (LSTM) RNNs \cite{Hochreiter:97lstm}. This and the later gated LSTM version \cite{Gers:2000nc} informed the first very deep FNNs called highway networks \cite{srivastava2015training}. LSTM RNNs are particularly well-suited for tasks involving credit assignment over thousands of steps, whereas similar highway networks were the earliest FNNs with hundreds of layers (previous FNNs had at most tens of layers). Following the same principle, the popular ResNet FNN architecture \cite{he2016deep} keeps the highway gates permanently open, allowing uninterrupted information flow from the first to the last layer. Residual connections \cite{Hochreiter:91} have become essential for many successful deep-learning architectures \cite{huang2017densely}, including graph neural networks with hundreds of layers \cite{li2019deepgcns, li2021training}.

\section{Preliminaries}\label{sec_Preliminaries}

\paragraph{Reinforcement Learning.}
RL is usually formalized as a Markov decision process (MDP) problem \citep{puterman2014markov}. 
An MDP comprises states $s \in \mathcal{S}$, actions $a \in \mathcal{A}$, a reward function $\mathcal{R}(s,a,s')$, and a transition function ${\cal T}(s'|s,a)$ that represents the likelihood of transitioning to the next state $s'$ from the current state and action $(s,a)$. We assume that the action space is finite and the state space is countable.
A policy $\pi(a|s)$ defines a probability distribution over actions for each state.
The value function $V^{\pi}(s)$ is defined as the expected discounted sum of rewards for following policy $\pi$ from state $s$, i.e., $V^\pi (s) \triangleq \E \left[ \sum_{t=0}^{\infty} \gamma^t \mathcal{R}(s_t,a_t,s_{t+1})  | s_0=s; \pi \right] $, where $\gamma \in [0,1)$ is a discount factor.
It is also convenient to define the action-value function $Q^{\pi}(s,a) \triangleq \sum_{s^{\prime}}{\mathcal{T} \left( s^{\prime}|s,a \right)}\left[ \mathcal{R} \left( s,a,s^{\prime} \right) +\gamma V^\pi\left( s^{\prime} \right) \right].$
The objective of RL is to find a policy yielding the maximum expected sum of rewards.
To achieve this, the optimal value function is defined as follows: $V^*(s)= \max_{\pi} V^\pi(s)$ and $Q^*(s,a)= \max_{\pi} Q^\pi(s,a)$, which allow us to construct an optimal policy $\pi^*(s) = \argmax_{a} Q^*(s,a)$ that satisfies $V^{\pi^*}(s) = V^{*}(s)  \forall s$.
The Bellman optimality operator and Bellman expectation operator are commonly used to obtain these value functions as follows:
\begin{fontsize}{8}{1}
\begin{align}
(\BOOperator V )(s) \triangleq &
\max_a \sum_{s^{\prime}}{\mathcal{T} \left( s^{\prime}|s,a \right)}\left[ \mathcal{R} \left( s,a,s^{\prime} \right) +\gamma V\left( s^{\prime} \right) \right] ,
 \label{eq_BOOperator}
\\
(\BEOperator V )(s) \triangleq &
\sum_a{\pi (a|s)}\sum_{s^{\prime}}{\mathcal{T} \left( s^{\prime}|s,a \right)}\left[ \mathcal{R} \left( s,a,s^{\prime} \right) +\gamma V\left( s^{\prime} \right) \right].
 \label{eq_BEOperator}
\end{align}
\end{fontsize}Iteratively applying $\BOOperator$ and $\BEOperator$ to any initial value function $V^{(0)}$ will result in the convergence to $V^*$ and $V^{\pi}$, respectively. The \emph{value iteration (VI)} algorithm is a concrete example of such a convergence, which iteratively applies the Bellman optimality operator as $V_{}^{\left( n+1 \right)} = \BOOperator V^{(n)}$.

\begin{figure}
    \def\height{0.85}
    \centering
    \subfloat[VIN/HighwayVIN]{
    \includegraphics[height=\height\linewidth]{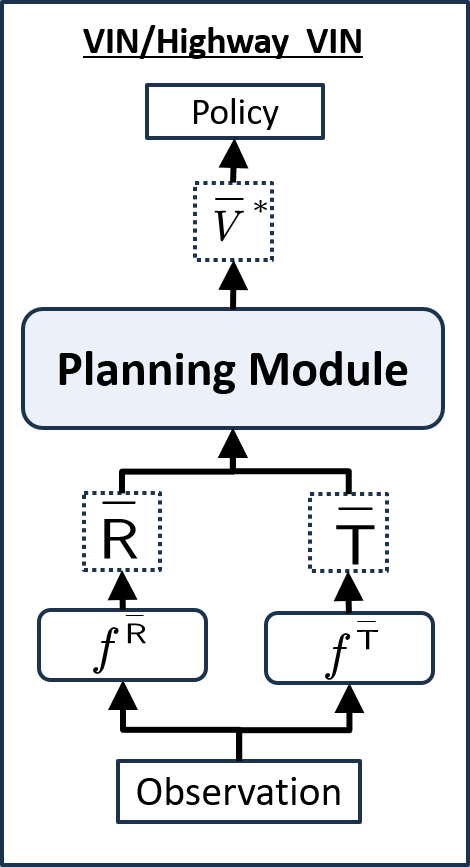}
    \label{fig_VIN_HighwayVIN}
    }
    \subfloat[Planning Module of VIN]{
    \includegraphics[height=\height\linewidth]{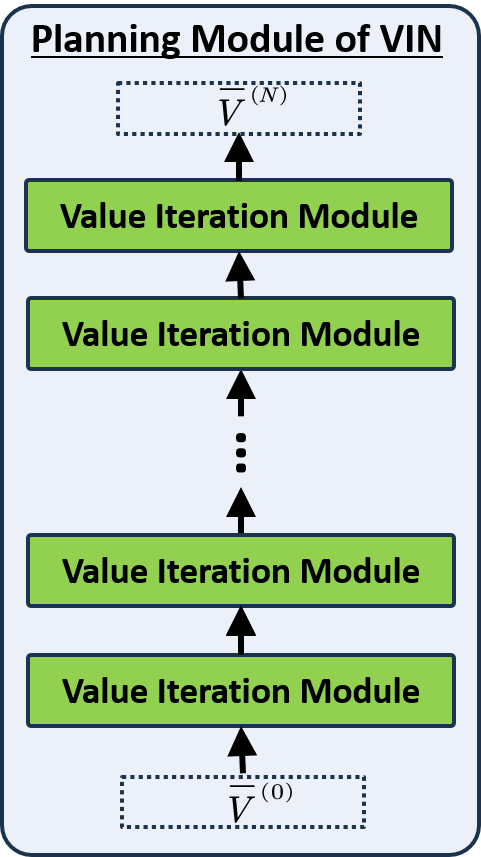}
    \label{fig_planning_VIN}
    }
    \caption{
        \subref{fig_VIN_HighwayVIN}: Architecture of VIN and highway VIN.
        \subref{fig_planning_VIN}: Architecture of the planning module of VIN, which includes $N$ layers of value iteration modules. The architecture of the value iteration module is detailed in \Cref{fig_VI_module}.
    }
    \label{fig_VIN_archiecture}
\end{figure}

\paragraph{Value Iteration Networks.}
VINs are NNs that integrate the process of planning into the learning architecture. 
VINs feature a ``planning module'', which approximates the VI process based on a learned latent MDP $\overline{\mathcal{M}}$. 
Below, we will use $\overline{\:\cdot\:}$ to denote all the terms associated with the latent MDP $\overline{\mathcal{M}}$.
VINs use learnable mapping functions to transit an observation $\phi(s)$ to a latent MDP by $\overline{\mathsf{R}}=f^{\overline{ \mathsf{R}} }( \phi(s) )$ and $\overline{\mathsf{T}}=f^{\overline{ \mathsf{T}} }( \phi(s))$. 
Then, it implements the VI update in \Cref{eq_BOOperator} using a \emph{Value Iteration module}, which applies a convolutional operation along with a max-pooling operation:
\begin{fontsize}{9.2}{1}
\begin{equation}\label{eq_Q_from_V}
\overline{Q}_{\overline{a},i,j}^{(n)}=\sum_{i^{\prime},j^{\prime}}{\left( \overline{\mathsf{T}}_{\overline{a},i^{\prime},j^{\prime}}^{}\overline{\mathsf{R}}_{i-i^{\prime},j-j^{\prime}}+\overline{\mathsf{T}}_{\overline{a},i^{\prime},j^{\prime}}^{}\overline{V}_{i-i^{\prime},j-j^{\prime}}^{(n-1)} \right)}
,
\end{equation}
\end{fontsize}
\begin{equation}\label{eq_V_from_Q}
\overline{V}_{i,j}^{(n)}=\max_{\overline{a}} \overline{Q}_{\overline{a},i,j}^{(n)}.
\end{equation}
Here, the indices $i,j$ correspond to the coordinates of the latent state, and $\overline{a}$ is the index of the action in the latent MDP $\overline{\mathcal{M}}$.
\Cref{eq_Q_from_V} sums over a matrix patch centered around position $(i,j)$.
\Cref{fig_VI_module} shows the computation process of the VI module.
By stacking the VI module for several layers, it approximates the optimal value function $\overline{V}^*$, which is then mapped to a policy applicable to the actual MDP $\mathcal{M}$.
\Cref{fig_VIN_archiecture} shows the VIN architecture.
As each component of the architecture is differentiable, VINs can be trained end-to-end.

\paragraph{Highway Value Iteration.}
Highway value iteration (highway VI) is an algorithm derived from the theory of highway RL \cite{wang2023highway}, which is a framework for improving the efficiency of long-term credit assignment.
This approach introduces a multi-step operator, which averages $n$-step bootstrapping values using various policies, termed  \emph{lookahead policies}, each executed for different $n$ time steps. Here, $n$ represents the \emph{lookahead depth}. This operator is formally defined as follows:
\begin{fontsize}{8.9}{1}
\begin{equation}\label{eq_highway_value_iteration}
\highwaySoftmaxOperator V
\triangleq
\SoftmaxPolicySoftmaxStep
{\color[RGB]{240, 0, 0} \max_{} }
\left\{ \left( \mathcal{B} ^{\pi} \right) ^{\circ \left( n-1 \right)}\mathcal{B} V,\mathcal{B} V \right\}.
\end{equation}
\end{fontsize}In this formula, $\policySet$ denotes the set of lookahead policies and $\stepSet$ denotes the set of lookahead depths. The softmax function, with a softmax temperature $\alpha$, is defined as:
$
\underset{x\in \mathcal{X}}{\softmax } 
f\left( x \right) 
\triangleq 
\sum_{x\in \mathcal{X}}
\frac
{{\exp \left( \alpha f\left( x \right) \right)}  }
{\sum_{x'\in \mathcal{X}}{\exp \left( \alpha f\left( x' \right) \right)}}
f\left( x \right).
$ Here, $(\cdot)^{\circ k}$ indicates the composition of operator $(\cdot)$ for $k$ times.
Based on this operator, the algorithm \emph{highway VI} iteratively updates the value function as $V^{(n+1)}= \highwaySoftmaxOperator V^{(n)} $.
Two critical aspects of highway VI are highlighted: 
\begin{remark}\label{remark_converging_to_optimal_V}
(Convergence to the Optimal Value Function)
The highway VI algorithm is proved to converge to the optimal value function $V^*$ 
regardless of the choice of lookahead policies $\policySet$, lookahead depths $\stepSet$, and softmax temperatures $\tilde\alpha$ and $\alpha$. 
For a detailed formal statement, please refer to their Theorem A.2 \cite{wang2023highway}.
\end{remark}
\begin{remark}\label{remark_wrongoperator_not_converge}
(Importance of the Maximization Operation)
The maximization operation, ${\color[RGB]{240, 0, 0} \max_{} }\left\{ \left( \mathcal{B} ^{\pi} \right) ^{\circ \left( n-1 \right)}\mathcal{B} V,\mathcal{B} V \right\}$, is crucial for ensuring convergence to $V^*$. Convergence is not guaranteed without this component.
For a detailed formal statement, please refer to their Theorem 1, 2 \cite{wang2023highway}.
\end{remark}

\section{Method}

\paragraph{Motivation.} 
NNs with increased depth have superior representational and generalization capabilities \cite{szegedy2014, ciresan:2011ijcai, ciresan:2012NN, telgarsky2016benefits}. 
Building on this knowledge, we propose that increasing the depth of VIN can considerably boost their long-term planning abilities in the context of RL.
This proposition is grounded in the intrinsic design of VINs, which includes a value iteration planning module.
A theoretical study (see Theorem 1.12, \cite{agarwal2019reinforcement}) indicates that increased iterations in this module can result in a more accurate estimation of the optimal value function, subsequently improving the policy performance.

\paragraph{Overview.}
The traditional VIN (\Cref{sec_Preliminaries}) propagates information layer-by-layer, based on the step-by-step approach of the VI process.
The proposed novel method, i.e., \emph{highway VINs}, enhances VINs by incorporating a distinct planning module inspired by highway VI.
As detailed in \Cref{sec_Preliminaries}, highway VI uses information from various policies and multiple steps ahead, forming a new VIN architecture that facilitates the information flow from various dimensions.
The planning module of highway VINs follows the computational process of highway VI in \cref{eq_highway_value_iteration}. 
Below, we detail the transition from the $n$-th activation $\V[n]$,  representative of the iterative process of the proposed planning module.

\begin{figure}[t]
    \def\height{1.2}
    \centering{
    \includegraphics[width=0.98\linewidth]{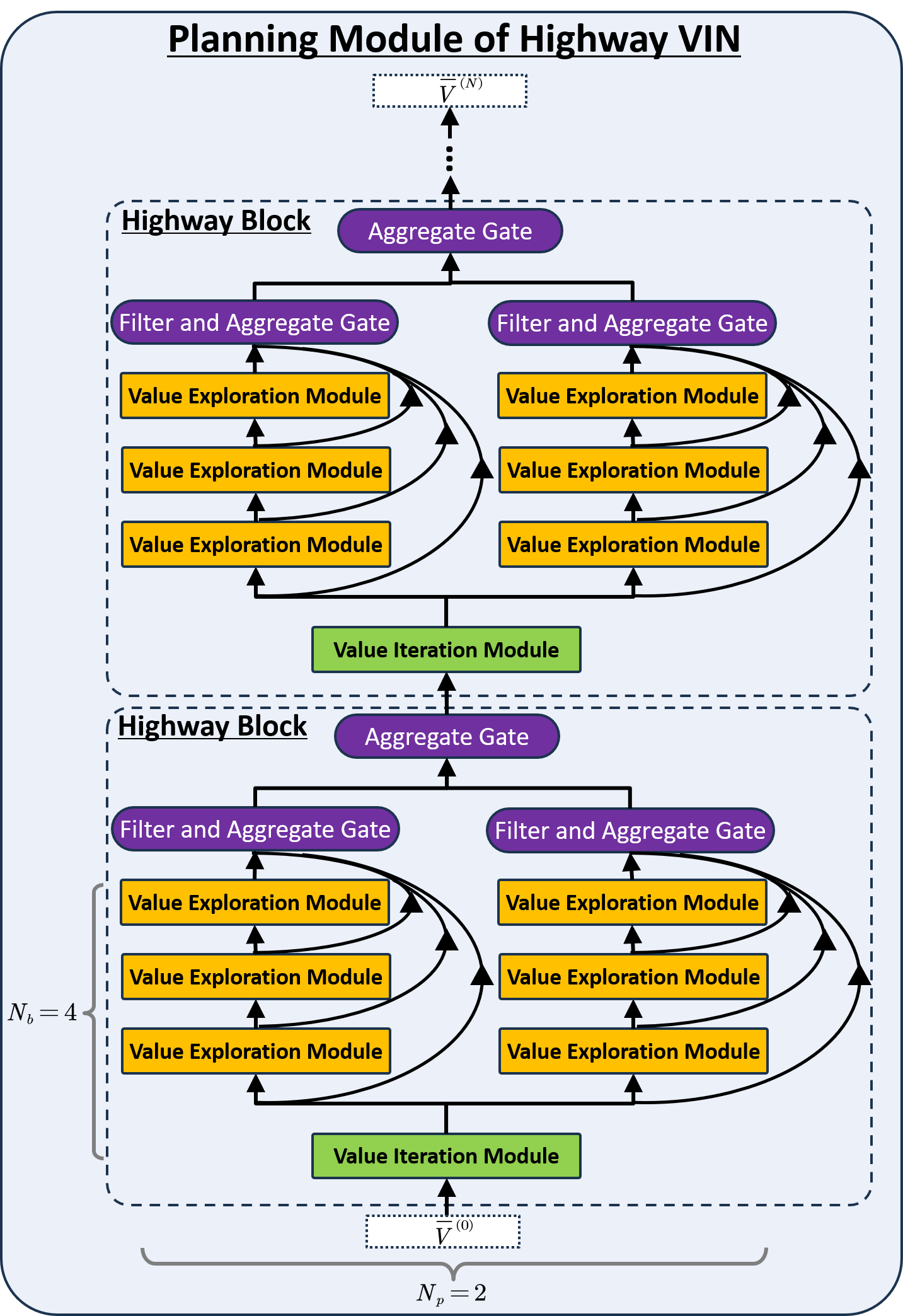}
    }
    \caption{  
        Planning module of highway VIN.
        Here, we demonstrate the planning module of highway VIN using a highway block of depth $N_b=4$ and incorporating $\policyCount=2$ embedded policies.
    }
    \label{planning_HighwayVIN}
\end{figure}

\textbf{First}, $\V[n]$ is fed into the value iteration module to generate a new activation $\V[n+1]$, as in VINs (\crefnop{eq_Q_from_V} and \ref{eq_V_from_Q}).
This step corresponds to the Bellman optimality operator $ \BOOperatorEmbed (\cdot) $ in highway VI (see \crefnop{eq_highway_value_iteration}).

\textbf{Then}, to facilitate information flow over spatial dimensions, we introduce a new \emph{value exploration (VE module) module}. Each VE module is equipped with an \emph{embedded policy} $\overline{\pi}$, defined on the latent MDP $\overline{\mathcal{M}}$ and determining the path of the information flow.
Conceptually, it corresponds to one application of the Bellman Expectation operator $ \BEOperatorEmbed (\cdot) $ in highway VI.

\textbf{Then}, to further facilitate spatial information flow in depth, we stack $\policyCount$ parallel VE modules for $N_{b}-1$ layers, corresponding to multiple compositions $({\BEOperatorEmbed})^{\circ (N_b-1)}(\cdot)$ in highway VI.
These stacked and parallel VE modules process the input $\overline{V}^{(n)}$ , leading to new activations $ \{ \vfVE{\overline{V}}^{(n+n_b)}_{\policyIndex}\}_{\policyIndex, n_b } $ for various indexes of the parallel modules $\policyIndex=1,\cdots, \policyCount $ and various depths $n_b=1, \cdots, N_{b}$, where the initial $\vfVE{\overline{V}}_{\policyIndex}^{(n+1)}={\overline{V}}^{(n+1)}$ for each $\policyIndex$.

\textbf{Finally}, the outputs from these parallel and stacked VE modules are combined using an \emph{aggregate gate} and a \emph{filter gate}. This combination forms a skip connection architecture, which eases the training of very deep NNs. These gates mirror the operations $\SoftmaxPolicySoftmaxStepNoSet
{\color[RGB]{240, 0, 0} \max_{} } \{\cdot\} $ in highway VI.

We term the above four procedures as a \emph{highway block}.
The planning module of highway VIN comprises $\blockCount$ such highway blocks.
\Cref{planning_HighwayVIN} overviews this planning module.
The subsequent sections detail the components of the VE module, filter gate, and aggregate Gate.

\subsection{Value Exploration Modules}\label{sec_value_exploration_module}

\begin{figure}
    \def\height{0.9}
    \centering{
    \subfloat[\tiny{Value Iteration Module}]{
    \includegraphics[height=\height\linewidth]{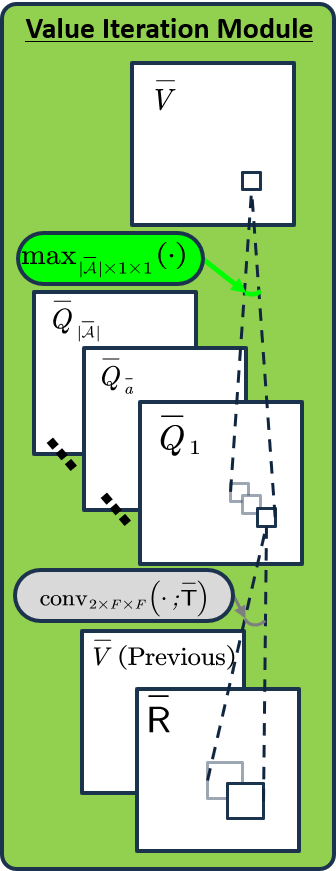}
        \label{fig_VI_module}
    }
    \subfloat[\tiny{Value Exploration Module}]{
    \includegraphics[height=\height\linewidth]{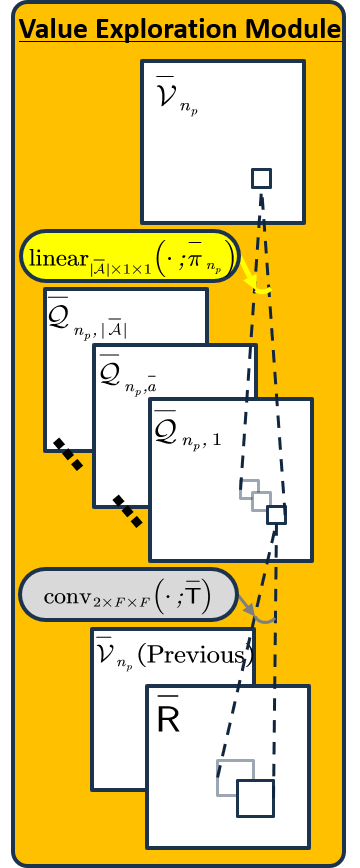}
        \label{fig_VE_module}  
    }
    }
    \caption{
        Architecture of the value iteration module and VE module, respectively.
        The operation $\max_{\overline{\mathcal{A}} \times 1 \times 1 }$ denotes a max operation over the action axis, as shown in \cref{eq_V_from_Q}.
        The operation $\mathrm{linear}_{\overline{\mathcal{A}} \times 1 \times 1 }$ represents a linear combination of the input Q matrix $\overline{Q}_{ \policyIndex }$ and the policy matrix $ \overline{\pi}_{\policyIndex} $ over the action axis, as shown in \cref{eq_V_from_Q__Highway}.
    }
    \label{fig:enter-label}
\end{figure}

The value iteration module in VINs greedily takes the largest Q value, as shown in \cref{eq_V_from_Q}. 
Consequently, this mechanism can result in a distinctive information flow for each layer, thereby channeling gradients toward certain specific neurons.
To facilitate the information and gradient flow across spatial dimensions, we introduce a new \emph{VE module}.
Each VE module is equipped with an \emph{embedded policy}, which determines the path of the information flow.
The VE module computes values according to Bellman expectation operator $ \BEOperatorEmbed (\cdot) $ in highway VI:

\begin{fontsize}{7.8}{1}
\begin{equation}\label{eq_Q__Highway}
\overline{\mathcal{Q} }_{\overline{\pi },\overline{a},i,j}^{(n+n_b)}=\sum_{i^{\prime},j^{\prime}}{\left( \overline{\mathsf{T}}_{\overline{a},i^{\prime},j^{\prime}}^{}\overline{\mathsf{R}}_{i-i^{\prime},j-j^{\prime}}+\overline{\mathsf{T}}_{\overline{a},i^{\prime},j^{\prime}}^{}\overline{\mathcal{V} }_{\overline{\pi },i-i^{\prime},j-j^{\prime}}^{(n+n_b-1)} \right)}
\end{equation}
\end{fontsize}

\begin{equation}\label{eq_V_from_Q__Highway}
\overline{\mathcal{V} }_{n_p,i,j}^{(n+n_b)}=\sum_{\overline{a}}{\overline{\pi }_{n_p,\overline{a},i,j}^{\left( n+n_b \right)}\overline{\mathcal{Q} }_{n_p,\overline{a},i,j}^{(n+n_b)}}
\end{equation} 
where
$
\policy \in \R ^{  |\mathcal{A}| \times m \times m  }
$ represents the $\policyIndex$-th {embedded policy} for $(n+n_b)$-th layer.
Here, the value functions are denoted as $\vfVE{\overline{V}}$ and $\vfVE{\overline{Q}}$ to distinguish them from the value functions $\overline{V}$ and $\overline{Q}$ of the VI module.
As implied by \cref{eq_V_from_Q__Highway}, instead of taking maximization over the actions as in the VI module (see \Crefnop{eq_V_from_Q}),
the proposed VE module takes expectation over actions based on the distribution of the embedded policy $\policy[][\policyIndex]$.
The computation process of the VE module is shown in \Cref{fig_VE_module}.

Note that, as stated in Remark \ref{remark_converging_to_optimal_V}, convergence in highway VI is assured regardless of the chosen lookahead policies. These policies correspond to embedded policies within several VE modules of highway VINs.
Generally, 
the embedded policy can be generated using a learnable mapping function 
$
\policy =f_{}^{\overline{\pi }}\left( \vfVE{\overline{Q}}_{\policyIndex}^{(n+ n_b)},\phi(s);W^{(n+ n_b)}_{\policyIndex} \right) 
$, where $W^{ (n+ n_b) }_{\policyIndex}$ are learnable parameters. 
Drawing inspiration from the dropout technique, which introduces stochasticity into the activations to increase robustness \cite{srivastava2014dropout, hanson1990, Hertz:91, baldi2013understanding},
we randomly generate multiple embedded policies as follows:
\begin{equation}\label{eq_sampled_action}
\policy[(n+n_b)][\policyIndex,\overline{a},i,j]
=\begin{cases}
	1, & \overline{a}=\widehat{\overline{a}}\sim P\left( \cdot ;\mathcal{\overline{Q}}_{\policyIndex,\cdot ,i,j}^{\left( n + n_b\right)},\epsilon \right)\\
	0, &\text{otherwise,}
\end{cases}
\end{equation}
where 
$\epsilon$ is the \emph{embedded exploration rate}, and $\widehat{\overline{a}}$ is a sampled action drawn from the $\epsilon$-greedy distribution $P\left( \cdot;\mathcal{\overline{Q}}_{\policyIndex,\cdot ,i,j}^{\left( n + n_b \right)},\epsilon \right) $, defined as:
$$
P\left( \overline{a};\mathcal{\overline{Q}}_{n_p,\cdot ,i,j}^{\left( n+ n_b \right)},\epsilon \right) =\begin{cases}
	1-\epsilon + \frac{\epsilon}{|\mathcal{ \overline{A} } |} ,&\overline{a}= \argmax \limits_{\overline{a}^{\prime}}\mathcal{\overline{Q}}_{n_p,\overline{a}^{\prime},i,j}^{\left( n+ n_b \right)}\\
	\frac{\epsilon}{|\mathcal{\overline{A}} |} ,& \mathrm{otherwise.}
\end{cases}
$$

During the training phase, the embedded policies are generated randomly for each iteration. In contrast, we adopt greedy policies during the evaluation phase, where trained models are applied in realistic environments, defined as follows:
$$
\overline{\pi}_{n_p,\overline{a},i,j}^{\left( n + n_b \right)}=\begin{cases}
	1,&		\overline{a}=\mathrm{arg}\max_{\overline{a}^{\prime}} \overline{\mathcal{Q} }_{n_p,\overline{a}^{\prime},i,j}^{\left( n + n_b \right)}\\
	0,&		\mathrm{otherwise.}\\
\end{cases}
$$
This equation indicates that, during the evaluation phase, VE modules function in the same way as VI modules.
In practice, we observe that incorporating this stochastic mechanism substantially improves robustness and is essential for achieving high performance.
Moreover, this stochasticity does not impact the convergence to the optimal value function. As stated by the theory of highway VI in \Cref{remark_converging_to_optimal_V}, convergence is guaranteed irrespective of any chosen embedded policies, even if they are fully stochastic.

\subsection{Aggregate and Filter Gates}\label{sec_aggregate}
We aim to integrate the proven efficacy of skip connections in training very deep NNs \cite{srivastava2015training, he2016deep}. However, theoretically validating this architecture within the RL framework and ensuring its compatibility with the stochasticity in the VE modules are challenging. The highway VI algorithm offers a guiding principle for this design.
By implementing the $\SoftmaxPolicySoftmaxStepNoSet
{\color[RGB]{240, 0, 0} \max_{} } \{\cdot\} $ operations of highway VI, the activations are aggregated as follows:
\begin{fontsize}{7.5}{1}
\begin{equation*}\label{eq_aggregation}
\overline{V}_{i,j}^{(n+N_b)}=\sum_{n_p=1}^{N_p}{\widetilde{\mathsf{A}}_{n_p,i,j}^{\left( n+N_b \right)}\underset{{\overline{V}^{\prime\prime}}_{n_p,i,j}^{\left( n+N_b \right)}}{\underbrace{\sum_{n_b=1}^{N_b}{\mathsf{A}_{n_p,i,j}^{\left( n+n_b \right)}\overset{{\overline{V}^{\prime}}_{n_p,i,j}^{\left( n+n_b \right)}}{\overbrace{{\color[RGB]{240, 0, 0} \max_{} }\left\{ \overline{\mathcal{V} }_{n_p,i,j}^{\left( n+n_b \right)},\overline{V}_{i,j}^{\left( n +1 \right)} \right\}. }}}}}}
\end{equation*}
\end{fontsize}Here, 
$\widetilde{ \mathsf A }_{\policyIndex,i,j}^{\left( n+N_b \right)}$
and 
${ \mathsf A }_{\policyIndex,i,j}^{\left( n+n_b \right)}$
are termed as the \emph{aggregate gate}, reflecting the degree to which activations contribute to the output, similar to the concept in highway networks \cite{srivastava2015training,srivastava2015highway}.
The aforementioned equation illustrates the aggregation of activations from various layers across multiple parallel VE modules, i.e.,
$\left\{ \vfVE{\overline{V}}_{\policyIndex}^{\left( n + n_b \right)} \right\} _{\policyIndex,n_b}$, for
$\policyIndex=1,\cdots ,\policyCount$ and 
$n_b=1,\cdots ,N_b $.
\Cref{planning_HighwayVIN} illustrates the information flow of this computation process.
Aggregate gates can be generally generated using mapping functions as follows:  
$
{ \mathsf A }_{\policyIndex}^{\left( n+n_b \right)}=f_{}^{ { \mathsf A } }\left( \left\{ \overline{{V}}_{\policyIndex}^{(n+n_b^{\prime} )} \right\} _{n_b^{\prime}},\phi (s);U_{\policyIndex}^{(n+n_b)} \right) 
$. Here, $U^{ (n+n_b) }_{\policyIndex}$ are learnable parameters. 
For simplicity and consistency with highway VI, we use softmax weights in the following form:
\begin{equation*}
\begin{aligned}
&
\widetilde{\mathsf{A}}_{n_p,i,j}^{\left( n+N_b \right)}=\frac{\exp \left( \alpha _{\widetilde{\mathsf{A}}}^{}{\overline{V}^{\prime\prime}}_{n_p,i,j}^{\left( n+N_b \right)} \right)}{\sum\limits_{n_{p}^{\prime}}{\exp \left( \alpha _{\widetilde{\mathsf{A}}}^{}{\overline{V}^{\prime\prime}}_{n_{p}^{\prime},i,j}^{\left( n+N_b \right)} \right)}}
,
\text{for various } \policyIndex,
\\
& 
\mathsf{A}_{n_p,i,j}^{\left( n+n_b \right)}=\frac{\exp \left( \alpha _{\mathsf{A}}^{}{\overline{V}^{\prime}}_{n_p,i,j}^{\left( n+n_b \right)} \right)}{\sum\limits_{n_{b}^{\prime}}{\exp \left( \alpha _{\mathsf{A}}^{}{\overline{V}^{\prime}}_{n_p,i,j}^{\left( n+n_{b}^{\prime} \right)} \right)}}
,
\text{for various } n_b,
\end{aligned}    
\end{equation*}
where $\alpha _{\widetilde{\mathsf{A}}}^{}$ and $\alpha _{\mathsf{A}}^{}$ are the softmax temperatures that vary for each highway block and learnable via backpropagation.

The maximization operation $\color{red}{\max}\{\cdot\}$, which we term \emph{filter gate}, compares the $(n+n_b)$-th activation $\vfVE{\overline{V}}_{\policyIndex}^{\left( n+n_b \right)}$ with the $(n+1)$-th one  $\overline{V}_{}^{\left( n+1 \right)}$, selecting the maximum value.
This filter gate is essential for discarding any activations
$\vfVE{\overline{V}}_{\policyIndex}^{\left( n+n_b \right)}$ that are lower than $  \overline{V}_{}^{\left( n+1 \right)} $, effectively filtering out explorations in VE modules that do not contribute to convergence.
Furthermore, as suggested in \Cref{remark_wrongoperator_not_converge}, this operation is crucial for ensuring convergence.

\subsection{Relation to Highway Networks}\label{sec_relation_with_existing}
Section \ref{sec_aggregate}  and \Cref{planning_HighwayVIN} illustrate the planning module of highway VIN, which features an architecture with skip connections similar to those found in established NNs such as highway networks \cite{srivastava2015training, srivastava2015highway} and their variants residual networks \cite{he2016deep} and DenseNets \cite{huang2017densely}.
A straightforward approach is to directly implement skip connections in VINs as follows: 
\begin{equation}
\overline{V}_{i,j}^{(n+N_b)}=\sum_{n_b=1}^{N_b}{\mathsf{A}_{i,j}^{\left( n+n_b \right)}\overline{V}_{i,j}^{\left( n+n_b \right)}}.
\end{equation}
Here, $\overline{V}_{i,j}^{\left( n+n_b \right)}$ is derived from the VI module (Eqs. \ref{eq_Q_from_V} and \ref{eq_V_from_Q}), using $\overline{V}_{i,j}^{\left( n+n_b-1 \right)}$ as the input.
While this method helps to address optimization challenges in training very deep VINs, it has not shown effectiveness in improving long-term planning capabilities (\Cref{table_success_rate} in \Cref{sec_experiment}). 
Highway VINs include two additional critical components: a VE module that improves the diversity of information and gradient flow, and an innovative filter gate designed to eliminate useless information generated by the VE modules.
This study also reveals the underlying connections between the highway RL and highway networks, which were initially proposed under different contexts and purposes but share fundamental similarities.

\begin{table*}[t]
\centering
\caption{ 
The success rates for each algorithm with various depths under 2D maze navigation tasks with different ranges of shortest path length.
Please also refer to Appendix \Cref{table_success_rate__all_depths} for the results of all the other depths.
\label{table_success_rate} 
}
\resizebox{\textwidth}{!}{
\begin{tabular}{c|l|c|c|c|l|c|c|c|}
\hline
Maze Size & \multicolumn{4}{c|}{$15\times 15$} & \multicolumn{4}{c|}{$25\times25$}\\ \hline
Shortest Path Length && $[1,30]$ & $[30,60]$ & $[60,100]$ && $[1,60]$ & $[60,130]$ & $[130,230]$ \\ \hline
\multirow{4}{*}{ \thead{VIN \\ \cite{tamar2016value}}} & $N=20$ & $99.83 \pm 0.11$ & $96.48 \pm 0.58$ & $63.03 \pm 3.20$ & $N=30$ & $98.84 \pm 0.16$ & $49.25 \pm 4.16$ & \phantom{9}$2.96 \pm 0.66$ \\
& $N=40$ & $99.79 \pm 0.10$ & $95.84 \pm 0.69$ & $76.16 \pm 1.87$ & $N=60$ & $96.47 \pm 1.33$ & $48.26 \pm 4.21$ & \phantom{9}$7.87 \pm 3.54$ \\
& $N=100$ & \phantom{9}$0.80 \pm 0.03$ & \phantom{9}$0.00 \pm 0.00$ & \phantom{9}$0.00 \pm 0.00$ & $N=150$ & \phantom{9}$0.22 \pm 0.08$ & \phantom{9}$0.00 \pm 0.00$ & \phantom{9}$0.00 \pm 0.00$ \\
& $N=200$ & \phantom{9}$0.56 \pm 0.00$ & \phantom{9}$0.00 \pm 0.00$ & \phantom{9}$0.00 \pm 0.00$ & $N=300$ & \phantom{9}$0.24 \pm 0.00$ & \phantom{9}$0.00 \pm 0.00$ & \phantom{9}$0.00 \pm 0.00$ \\\hline
\multirow{4}{*}{\thead{GPPN \\ \cite{lee2018gated}}} & $N=20$ & $99.98 \pm 0.01$ & $92.68 \pm 1.07$ & $51.12 \pm 5.00$ & $N=30$ & $98.98 \pm 0.25$ & $25.98 \pm 5.78$ & \phantom{9}$2.76 \pm 1.68$ \\
& $N=40$ & ${\color{blue}\mathbf{99.99 \pm 0.01}}$ & $96.16 \pm 3.56$ & $65.17 \pm 12.4$ & $N=60$ & ${\color{blue}\mathbf{99.09 \pm 0.19}}$ & $28.87 \pm 1.47$ & \phantom{9}$1.32 \pm 0.55$ \\
& $N=100$ & $99.95 \pm 0.05$ & $93.34 \pm 4.16$ & $60.57 \pm 13.6$ & $N=150$ & $98.51 \pm 0.31$ & $21.62 \pm 3.50$ & \phantom{9}$0.73 \pm 0.68$ \\
& $N=200$ & $99.98 \pm 0.01$ & $92.79 \pm 1.28$ & $50.88 \pm 3.59$ & $N=300$ & $95.38 \pm 2.01$ & \phantom{9}$6.29 \pm 4.35$ & \phantom{9}$0.02 \pm 0.03$ \\\hline
\multirow{3}{*}{\thead{Highway network \\ \cite{srivastava2015training}}} & $N=40$ & $99.65 \pm 0.17$ & $96.04 \pm 0.63$ & $75.86 \pm 10.0$ & $N=60$ & $97.93 \pm 0.56$ & $62.95 \pm 8.79$ & $17.46 \pm 5.45$ \\
& $N=100$ & $99.36 \pm 0.32$ & $91.11 \pm 2.64$ & $60.32 \pm 8.87$ & $N=150$ & $85.42 \pm 4.20$ & $12.55 \pm 3.89$ & \phantom{9}$0.35 \pm 0.23$ \\
& $N=200$ & \phantom{9}$0.73 \pm 0.12$ & \phantom{9}$0.00 \pm 0.00$ & \phantom{9}$0.00 \pm 0.00$ & $N=300$ & \phantom{9}$0.24 \pm 0.00$ & \phantom{9}$0.00 \pm 0.00$ & \phantom{9}$0.00 \pm 0.00$ \\\hline
\multirow{3}{*}{\thead{Highway VIN \\ (ours)} } & $N=40$ & $99.77 \pm 0.09$ & $98.83 \pm 0.25$ & $90.00 \pm 2.12$ & $N=60$ & $97.87 \pm 0.60$ & $77.02 \pm 6.30$ & $20.68 \pm 9.89$ \\
& $N=100$ & $99.93 \pm 0.03$ & ${\color{blue}\mathbf{99.52 \pm 0.12}}$ & ${\color{blue}\mathbf{98.61 \pm 0.66}}$ & $N=150$ & $97.77 \pm 0.48$ & $89.56 \pm 0.95$ & $75.42 \pm 10.1$ \\
& $N=200$ & $99.94 \pm 0.01$ & $99.13 \pm 0.12$ & $98.20 \pm 1.75$ & $N=300$ & $98.73 \pm 0.50$ & ${\color{blue}\mathbf{92.28 \pm 3.50}}$ & ${\color{blue}\mathbf{90.06 \pm 3.13}}$ \\\hline
\end{tabular}
}
\end{table*}

\section{Experiments}\label{sec_experiment}

We conduct a series of experiments to evaluate how highway VINs can improve the long-term planning capabilities of VINs for complex tasks. 
We also explore the significance of each component within the highway VINs.

\subsection{2D Maze Navigation}
We evaluate the algorithms 
on 2D maze navigation tasks of various sizes $m\times m$, specifically $15 \times 15$ and $25 \times 25$.
In these tasks, the agent can move forward, turn 90 degrees left or right, and has four orientations. 
We follow the experimental setup described in the paper on GPPN \cite{lee2018gated}.
We train the models for $30$ epochs using imitation learning on a labeled training dataset, 
select the best model based on validation dataset performance, 
and test it on a separate test dataset. Each dataset contains numerous planning tasks, each involving a maze with a start position, an image of the $m \times m$ map, and a goal position represented by a $4\times m \times m$ matrix (where $4$ corresponds to the four orientations). 
These datasets involve tasks requiring planning over hundreds of steps to reach the goal.
For more detailed information about the dataset, please refer to \Cref{sec_experiment_detail_app}.

We measure the planning abilities of an agent based on the \emph{success rate (SR)}, which is defined as the ratio of the number of successfully completed tasks to the total number of planning tasks.
The agent is considered to succeed in a task if it generates a path from the start position to the goal position within a limited number of steps.	
To assess the planning ability of the algorithms across different scales, we evaluate them on navigation tasks with varying \emph{shortest path lengths (SPLs)} from start to goal. These lengths are calculated in advance using Dijkstra's algorithm with access to the maze's underlying structure.
Tasks with longer SPLs typically demand greater long-term planning capabilities. We follow the GPPN paper's setting, evaluating each algorithm with $3$ random seeds. 
We report the mean and standard deviation on the test dataset.

Subsequently, we compare the highway VINs against several advanced NNs for planning tasks.
The baseline model is the original VIN \cite{tamar2016value}.
We also compare highway VINs against GPPNs \cite{lee2018gated}, which improve the training stability of VINs using gated recurrent operators such as LSTM updates and the highway networks \cite{srivastava2015training}, which incorporate skip connections for training of very deep NNs (adapted here for VINs, \Cref{sec_relation_with_existing}).
We follow the hyperparameter settings listed in the paper on GPPNs \cite{lee2018gated}.
To ensure a fair comparison, we set the number of parallel VE modules of highway VINs to $\policyCount = 1$ unless stated otherwise.
The embedded exploration rate is set to $\epsilon=1$ (defined in \Crefnop{eq_sampled_action}). 
Note that this setting does not result in a sub-optimal solution of the latent value function due to the filter gate, which excludes actions detrimental to convergence.
The baseline uses a $20$-layer VIN and GPPN for the $15 \times 15$ Maze, and a $30$-layer VIN and GPPN for the $25 \times 25$ Maze.	
For both highway networks and highway VINs, we set $\blockCount=20$ highway blocks for the $15 \times 15$ maze and $\blockCount=30$ for the $25 \times 25$ maze.	
We set various highway block depths $\blockDepth \geq 2$, which yields various total depths: $N=\blockDepth * \blockCount$, specifically $N\in\{40,60,80,120,160,200\}$ for the $15\times 15$ maze and $N\in\{60, 90, 120, 150,180, 240, 300\}$ for the $25 \times 25$ maze.	
The baselines VIN and GPPN are also tested using the same depths.

\begin{figure}[t]
    \centering
    {\includegraphics[width=0.45\linewidth]{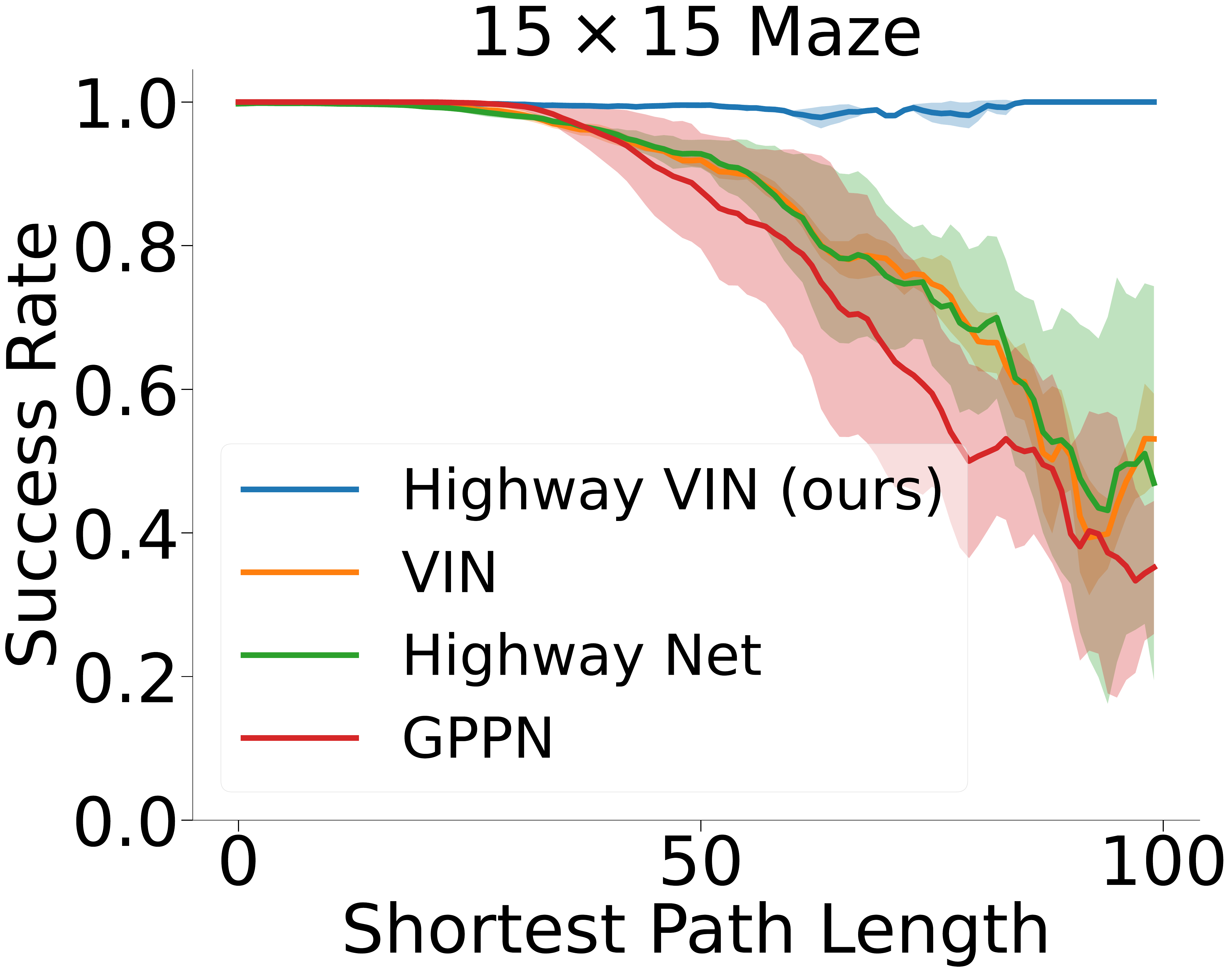}}
    {\includegraphics[width=0.45\linewidth]{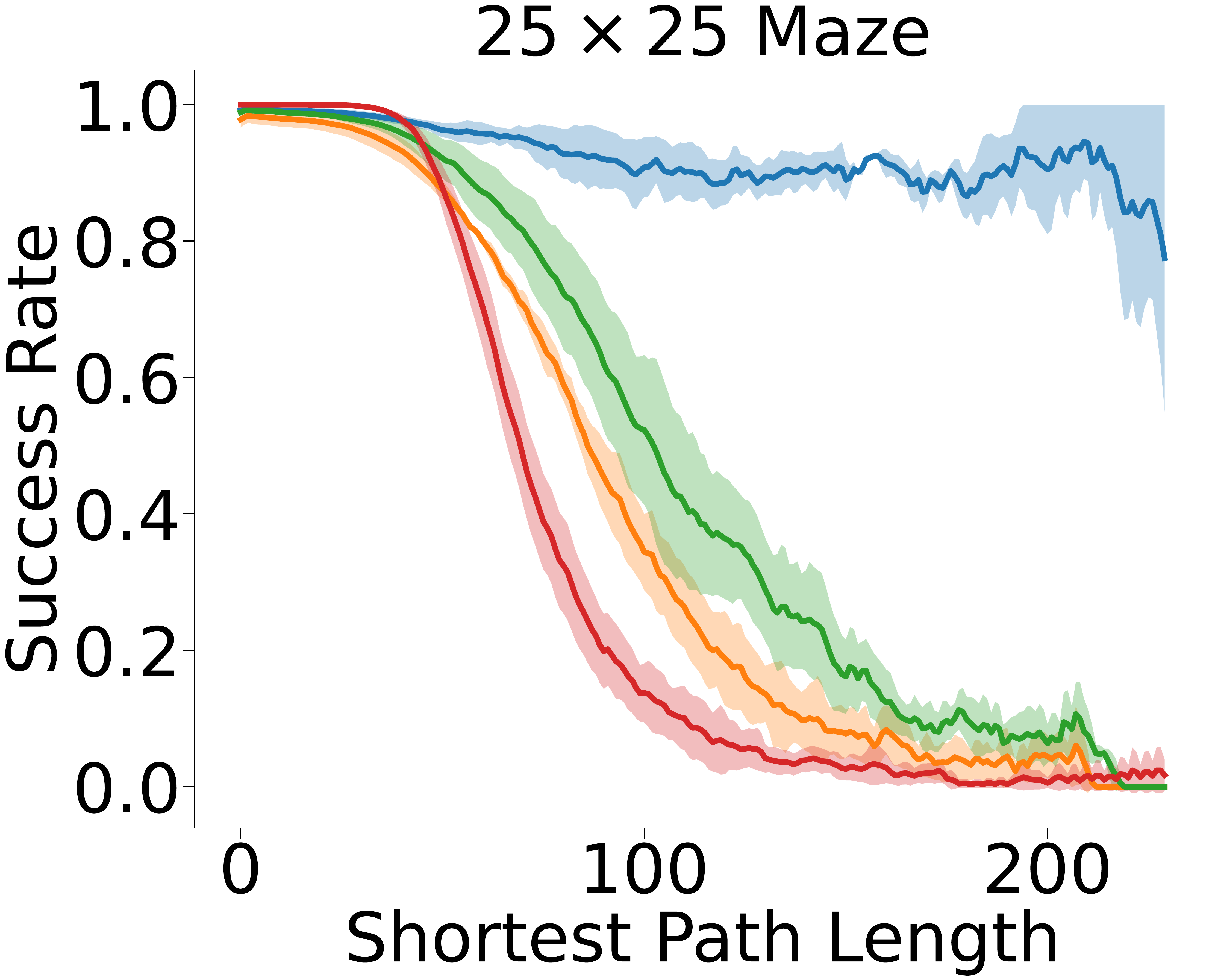}}
    \caption{
    Success rates of the algorithms are presented as a function of varying shortest path length.
    For each algorithm, the optimal result from a range of depths is selected. For a comprehensive view of the results across all depths, please see \Cref{fig__success_rate__algorithm__all_depths} in the Appendix.
    }
    \label{fig__success_rate__algorithm__best_depth}
\end{figure}

\begin{figure}[t]
    \centering
    \def\heightALAKEKW{0.28}
        \subfloat[$25 \times 25$ Maze \label{fig_maze} ]{
            \includegraphics[height=\heightALAKEKW\linewidth]{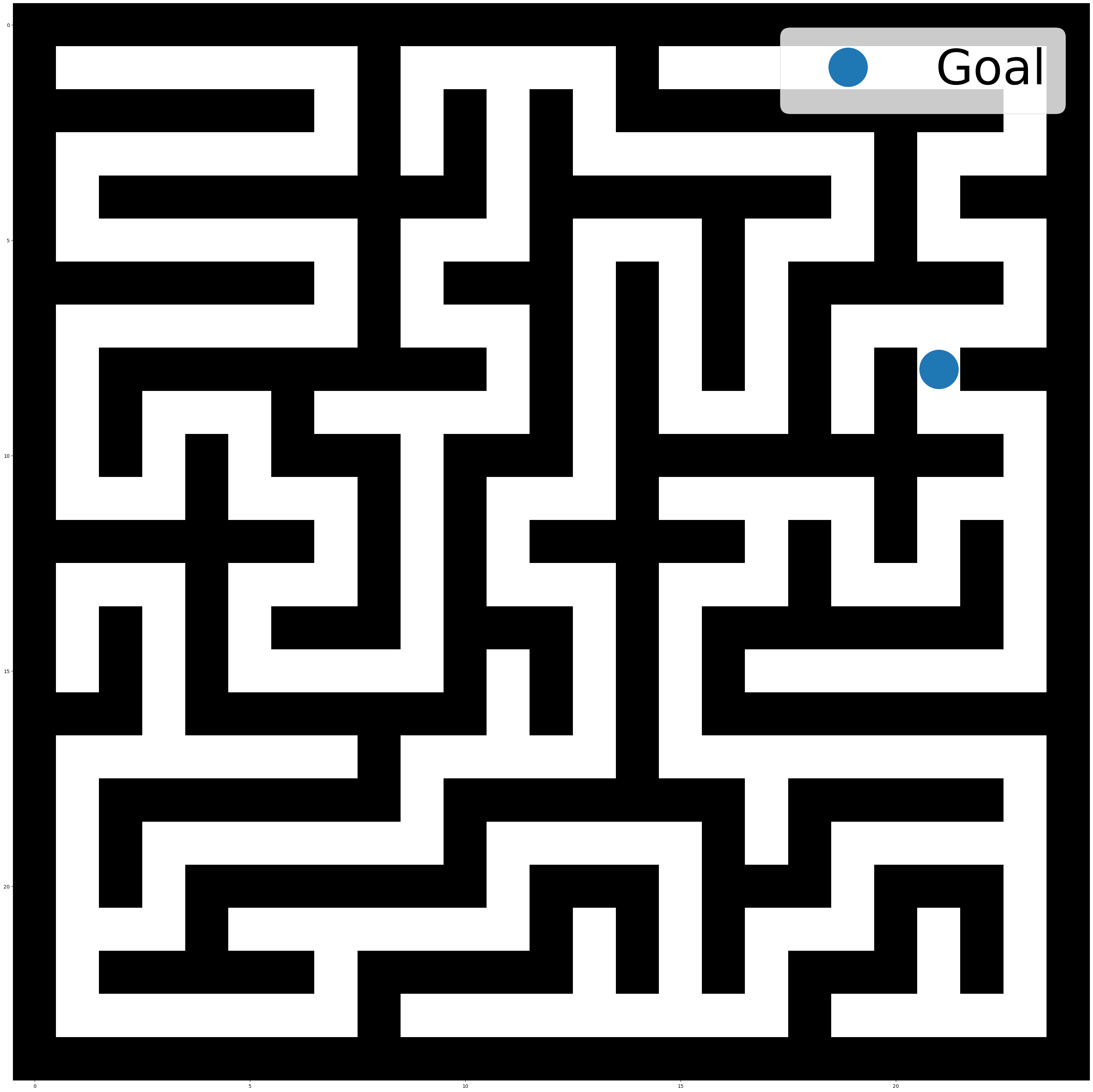} 
        }
        \subfloat[VIN \label{fig_heatmap_VIN} ]{
            \includegraphics[height=\heightALAKEKW\linewidth]{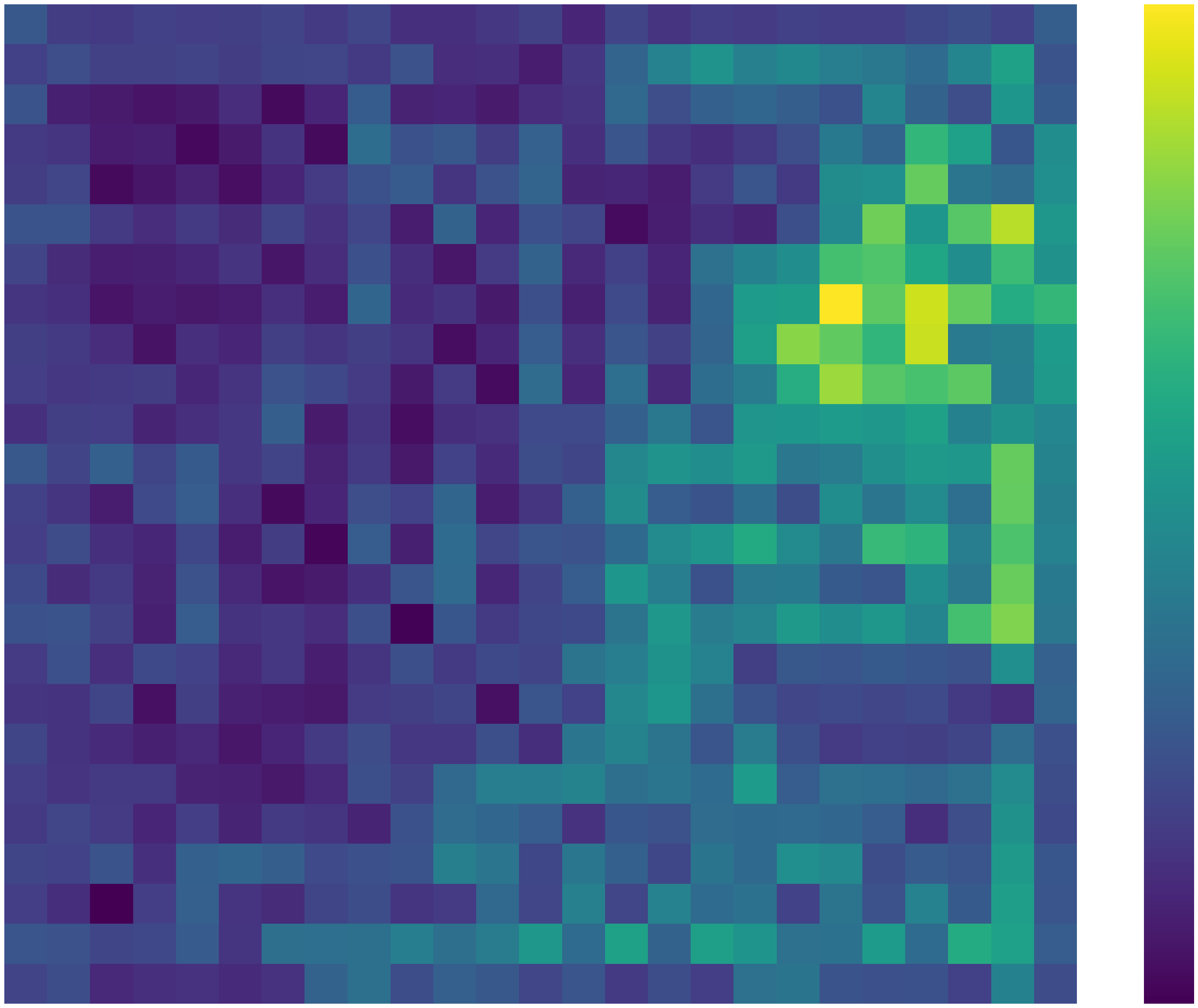} 
        }
        \subfloat[Highway VIN \label{fig_heatmap_HighwayVIN}]{
            \includegraphics[height=\heightALAKEKW\linewidth]{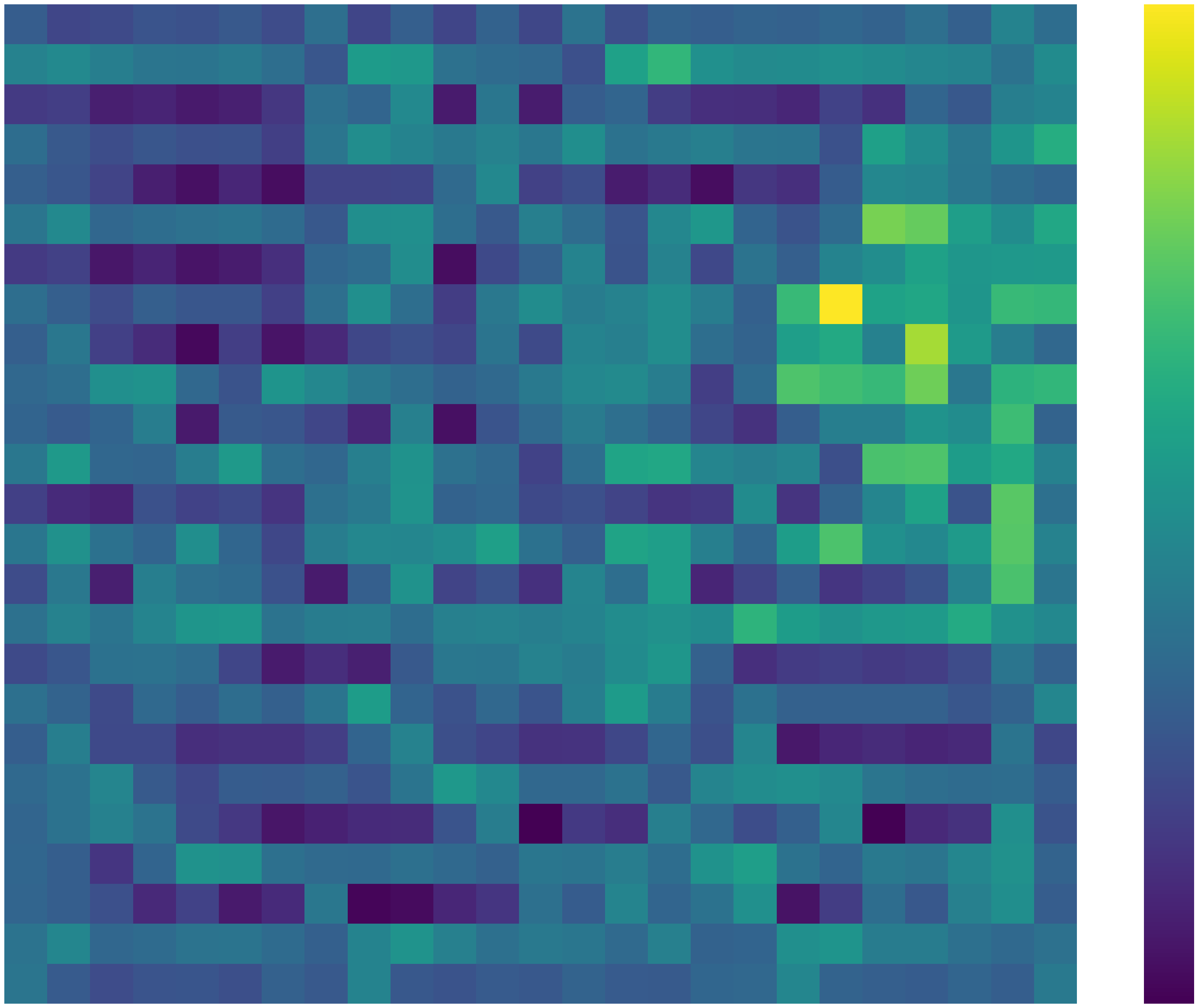}
        }
    \caption{
    \subref{fig_maze} An example of $25 \times 25$ Maze. 
    \subref{fig_heatmap_VIN} and \subref{fig_heatmap_HighwayVIN} learned feature maps of VIN and highway VIN, respectively.
    }
    \label{fig_heatmap}
\end{figure}

\paragraph{Peformance with the Best NN Depth.}
\Cref{fig__success_rate__algorithm__best_depth} shows the SRs of various algorithms under tasks with different SPLs. 
To ensure a fair comparison, we select the best results from varying NN depths for each algorithm as each algorithm may perform optimally at different depths (please refer to \Cref{fig__success_rate__algorithm__all_depths} in Appendix for a comprehensive view of the results across all depths).	
The results demonstrate that the SRs for all compared methods considerably decrease with increasing SPLs. Remarkably, when the SPL exceeds 200 in the $25 \times 25$ Maze, the SRs of all methods nearly drop to \textbf{0\%}.
In contrast, highway VIN maintains an impressive \textbf{98\%} SR with an SPL of $100$ in the $15 \times 15$ Maze and \textbf{90\%} SR with an SPL of $200$ in the $25 \times 25 $ Maze.

Additionally, \Cref{fig_heatmap} shows the feature map of the VIN and highway VIN, which can conceptually be understood as the learned value function.
The figure reveals that the learned values of highway VIN for states distant from the goal are larger than those of VIN, implying that highway VIN learns an effective value function for long-term planning.

\paragraph{Peformance with Various Depths.}
\Cref{table_success_rate} shows the SRs of each algorithm across various depths and tasks with different SPL ranges (see \Cref{table_success_rate__all_depths} in the Appendix for results across all depths).
The table shows that the highway VINs generally perform better with increased depth. Notably, we observe a \textbf{+69.38\%} improvement in the SR of the $25 \times 25$ maze with an SPL range of $[130, 230]$ as the depth $N$ increases from $60$ to $300$.

The performance of the VIN decreases with increasing depth. Specifically, the SR of VINs drops to nearly \textbf{0\%} for all tasks at a depth of $N>150$.
The highway network, in contrast, maintains its performance even at greater depths. However, an increase in depth does not considerably improve its long-term planning capabilities. We hypothesize that integrating the skip connections of highway networks into VINs does not introduce additional architectural inductive bias beneficial for planning.

The GPPN effectively mitigates the challenges of training very deep models and performs robustly across various depths. In particular, the GPPN excels in tasks with short SPLs, achieving a \textbf{99.09\%} SR on a $25 \times 25$ maze with an SPL range of $[1,30]$.	
However, the GPPN does not show a notable improvement in long-term planning capabilities with an increased depth. For instance, the  SR of the GPPN drops to less than \textbf{3\%} on a $25 \times 25$ maze with an SPL range of $[130,230]$.
This might be because the GPPN, as a black box method with less inductive bias towards planning, is more suited to learning patterns for short-term planning tasks rather than those requiring long-term planning skills.

\subsection{Ablation Study}
Several ablation studies were conducted to evaluate: 
1) the effectiveness of the filter gate (\Cref{sec_aggregate});
2) the impact of the VE module (\Cref{sec_value_exploration_module}). 
We also evaluate the influence of the number of parallel VE modules in \Cref{sec_number_parrallel_VE_modules}.
In the highway VIN experiments, the default hyperparameters include an exploration rate $\epsilon$ of 1, a single parallel VE module ($\policyCount=1$), and depth configurations of 200 for the $15\times15$ mazes and 300 for the $25\times25$ mazes.

\begin{figure}[h]
    \centering
    \begin{tabular}{cc}
        \includegraphics[width=0.45\linewidth]{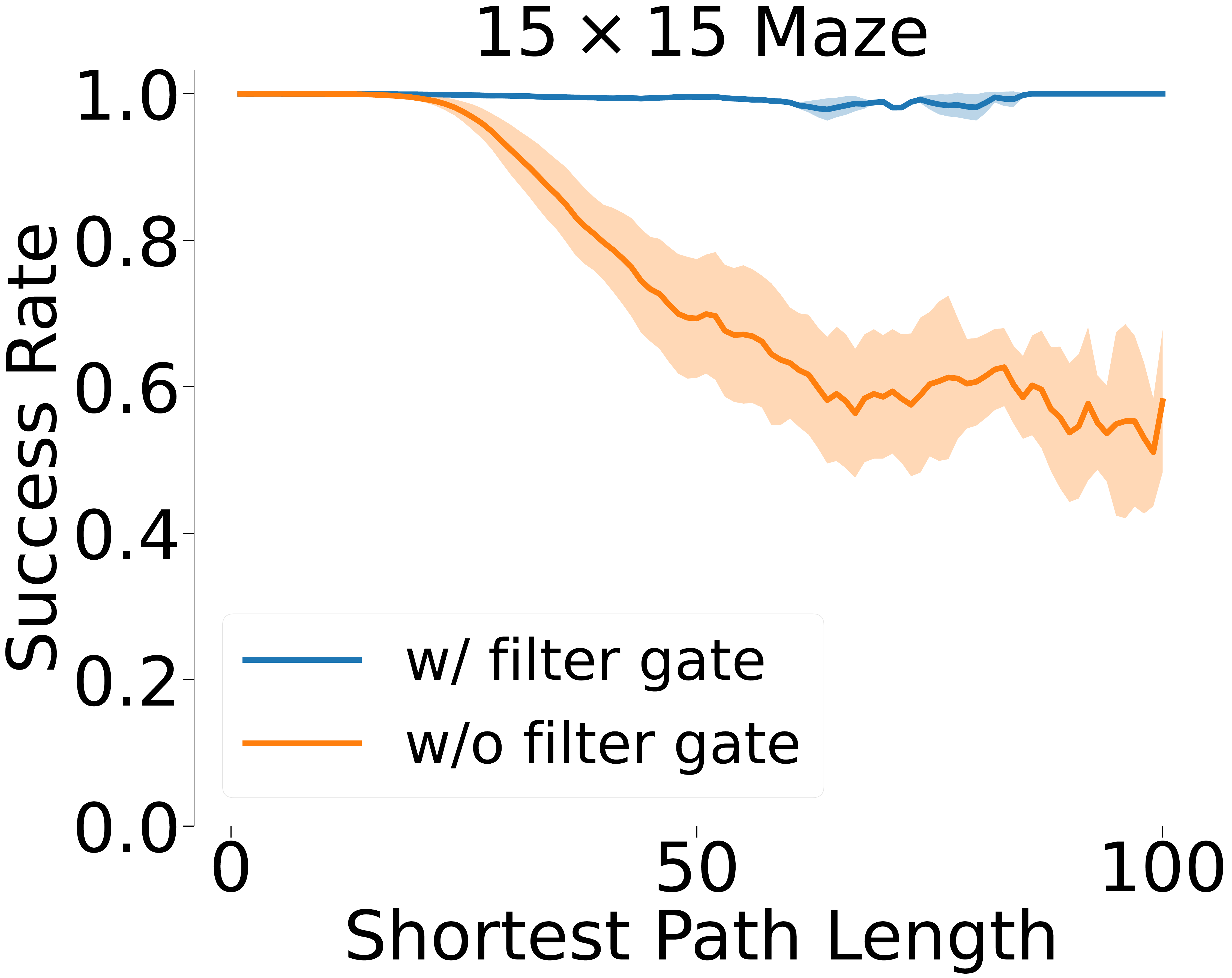} & \includegraphics[width=0.45\linewidth]{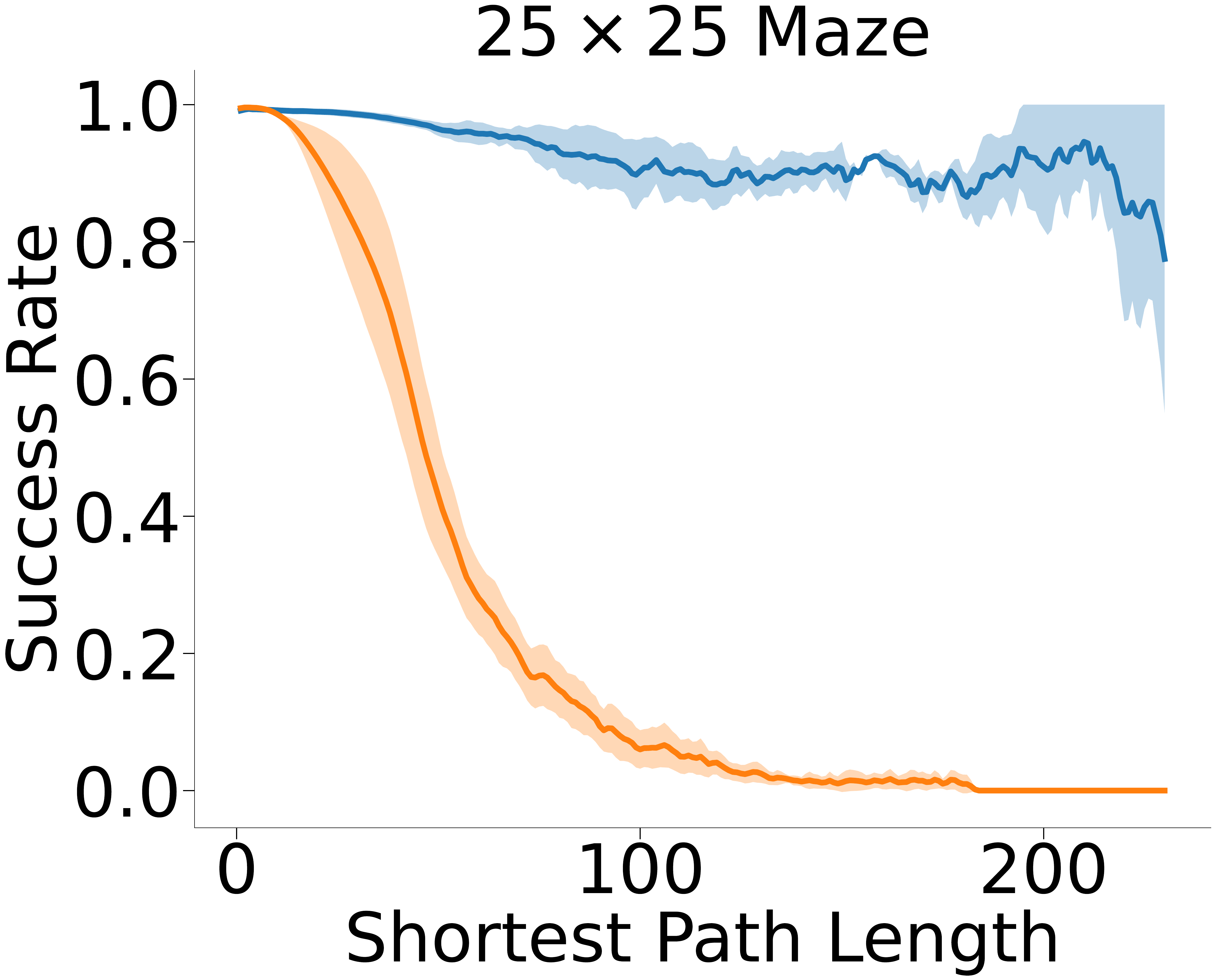}
    \end{tabular}
    \caption{
    Success rates of highway VIN with and without the filter gate.
    }
    \label{fig_filter_gate}
\end{figure}

\paragraph{Filter Gate.} The SRs of highway VINs, with and without the filter gate, are shown in \Cref{fig_filter_gate}.
The performance considerably decreases when the filter gate is absent.	
This is because, without a filter gate, highway VINs could easily suffer the adverse effects of exploration in VE modules, which could prevent convergence.

\begin{figure}[h]
    \centering
    \begin{tabular}{cc}
        \includegraphics[width=0.45\linewidth]{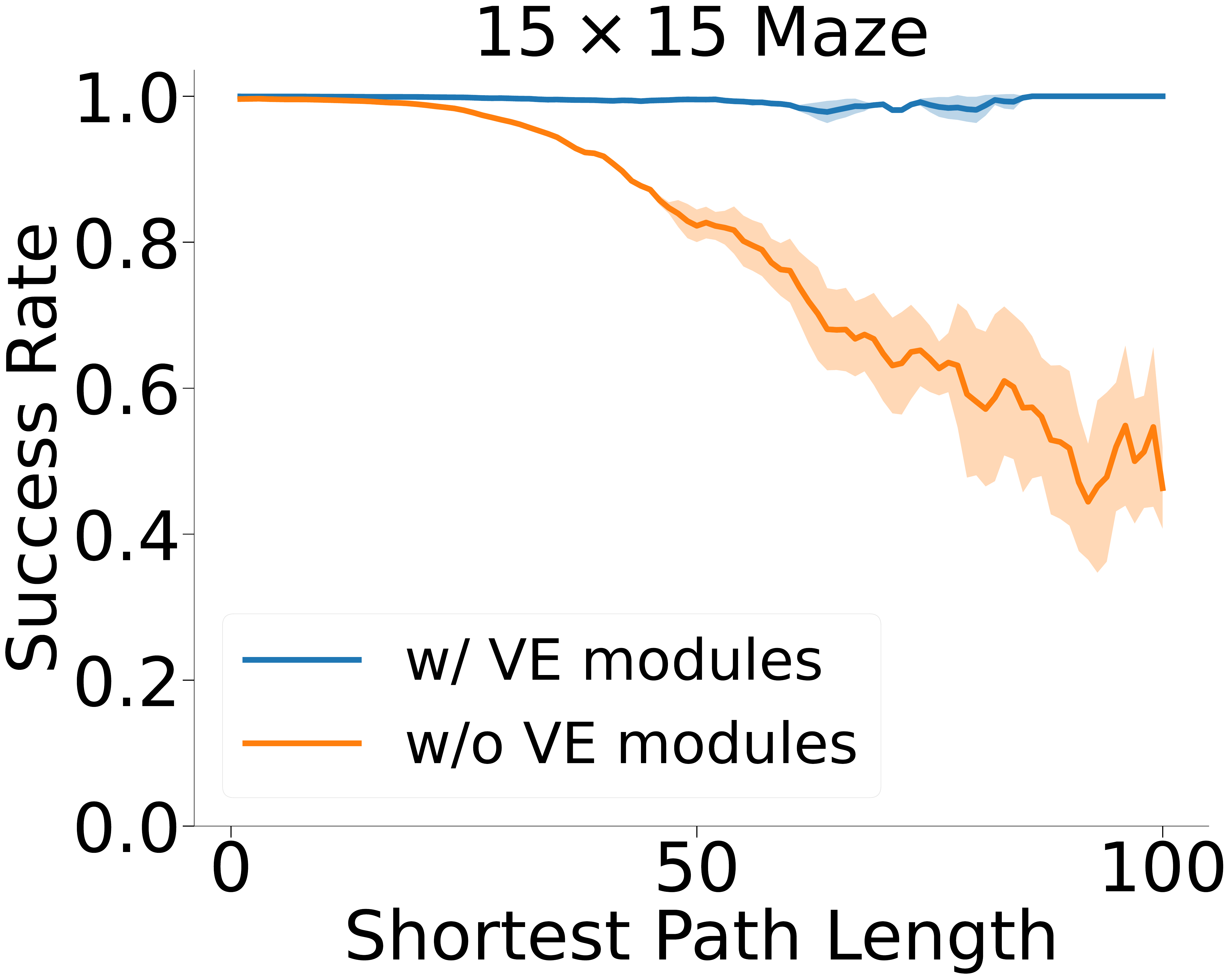} & \includegraphics[width=0.45\linewidth]{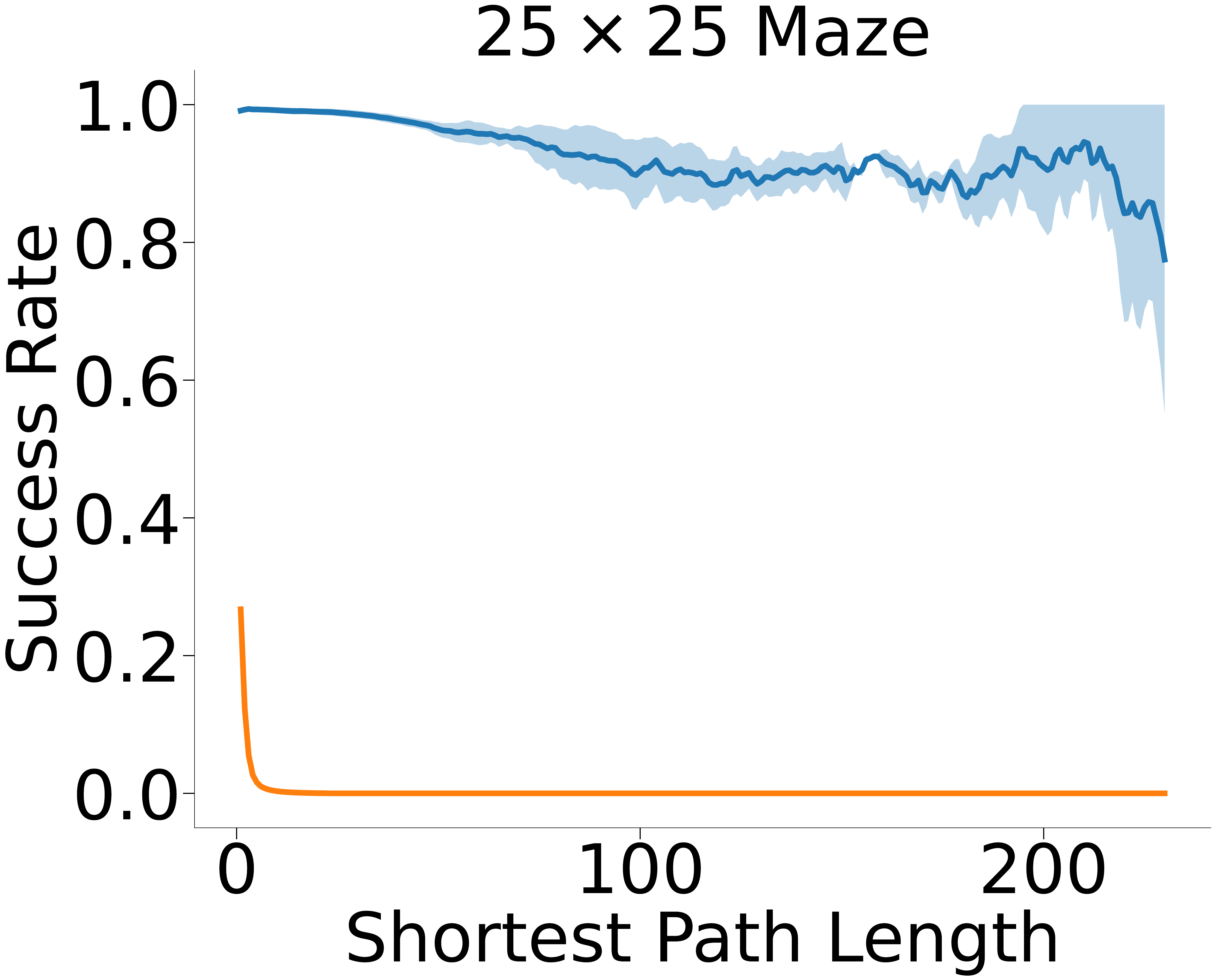}
    \end{tabular}
    \caption{
    Success rates of highway VIN with and without the VE modules,
    }
    \label{fig_stochasticity}
\end{figure}

\paragraph{VE Modules.} 
We also evaluate a variant of highway VIN without the VE module (referred to as \emph{w/o VE modules}), which means that all VE modules are replaced with VI modules.
\Cref{fig_stochasticity} shows the SRs for the variants with and without the VE module.
Without VE modules, the performance of highway VINs notably diminishes in the $25 \times 25$ maze.
In comparison, the variant equipped with the VE modules performs much better. 
This improvement is likely due to the use of diverse latent actions, which could facilitate information flow among various neurons in the NN. 
\Cref{table_entropy} lists the entropy of the selected latent actions of VIN and highway VIN.
For VIN, it always selects the action that leads to the maximum Q value.
Although the selected latent actions generally vary for each layer because the Q values dynamically change for each layer,	they will converge to the same actions when the Q values converge. 
Therefore, the diversity of the selected actions for VIN is much more limited. Instead, in the proposed highway VIN, we employ the value exploration module to maintain diversity.

\begin{table}[h]
    \centering
\caption{
    The entropy of the selected latent actions of VIN and highway VIN with various depths. The entropy is computed by $\sum_{\latentA \in \latentASpace  } {\left[ -p( \latentA ) \log p(\latentA) \right] }  $, where $p(\latentA)=\frac{ \mathrm{cnt}(\latentA) }{  \sum_{\latentA \in \latentASpace} {\mathrm{cnt} {(\latentA) } } }$ and $\mathrm{cnt}(\latentA)$ is the number of selected latent actions equals to $\latentA$ over the latent actions across all hidden layers.
}
    \resizebox{\columnwidth}{!}{
    \begin{tabular}{|c|c|c|c|c|c|c|c|}
        \hline
         Maze Size & \multicolumn{3}{c|}{$15\times 15$} & \multicolumn{3}{c|}{$25\times 25$}\\\hline
         Depth &  $40$ & $100$ & $200$ & $60$ & $150$ & $300$\\\hline
         VIN & 0.51 & 0.14 & 0.00 & 0.63 & 0.07 & 0.00 \\ \hline
Highway VIN (ours) & 2.04 & 3.53 & 4.17  & 2.27 & 3.77 & 4.35 \\ \hline
    \end{tabular}
    }
    \label{table_entropy}
\end{table}

\subsection{3D ViZDoom Navigation}
We evaluate the proposed approach in 3D ViZDoom environments \citep{Wydmuch2019ViZDoom}.
Following the experimental setting of the GPPN paper \citep{lee2018gated}, the input to the model consists of RGB images capturing the first-person view, rather than the top-down 2D maze. Based on these observations, a CNN is employed to predict the maze map, which is then inputted into the planning model (such as VIN or highway VIN), using an architecture and hyperparameters similar to the 2D maze setup.
We select the optimal depth from $N=40, 100, 200$ for each algorithm.
Highway VIN performs optimally at depth $100$, while other methods perform best at depth $40$.
\cref{table_SR_3D_ViZDoom} shows the SRs in $15 \times 15$ 3D ViZDoom maze navigation.
Our highway VINs excel in tasks with SPLs exceeding 30.

\begin{table}[h]
    \centering
    \caption{
Success rates of each algorithm with various depths under 3D ViZDoom maze navigation tasks with different ranges of SPLs.
        \label{table_SR_3D_ViZDoom}
    }
    \newcommand{\bestResult}[1]{{\color{blue}{\mathbf{#1}}}}
    \resizebox{\linewidth}{!}{
    \begin{tabular}{cccc}
        \toprule
        & $[1,30]$ & $[30,60]$ & [60,100] \\
        \midrule
        VIN  & $98.57\pm1.54$ & $92.03\pm2.03$ & $69.37\pm2.63$\\
        GPPN  & $ \bestResult{99.91\pm 0.10} $ & $89.95\pm9.21$ & $44.42\pm8.14$ \\
        Highway network & $97.82\pm1.01$ & $91.99\pm2.54$ & $63.78\pm11.49$ \\
        Highway VIN (ours)  & $99.43 \pm 0.18$ & $\bestResult{98.70 \pm 0.38}$ & $\bestResult{96.98\pm 1.20}$ \\
        \bottomrule
    \end{tabular}
    }
\end{table}

\subsection{Computational Complexity}

The proposed approach introduces a minimal number of additional parameters to the existing VIN architecture, specifically the softmax temperatures for each highway block denoted as $\{ (\alpha_{\widetilde{A}}^{(\blockIndex)},\alpha_{A}^{(\blockIndex)}) \}_{\blockIndex=1}^{ \blockCount }$.
Note that $\blockCount$ indicates the total number of highway blocks, which are set to $\blockCount=20$ for the $15\times 15$ maze and $\blockCount=30$ for the $25\times 25 $ maze.
Additionally, the following table details the GPU memory consumption and training duration for each method when employing $300$ layers on NVIDIA A100 GPUs.	

\resizebox{\linewidth}{!}{
\begin{tabular}{c|c|c|c|c}
\hline
              & VIN & GPPN & Highway network & Highway VIN (ours) \\
\hline
GPU Memory    &   3.1G  &   103.0G   &      3.3G       &          15.0G          \\
\hline
Training Time &   7.5 hours  &  6.5 hours    &    7.7 hours         &     9.0    hours          
\\
\hline 
\end{tabular}
}

\section{Conclusions}
This paper presents a general framework based on highway RL to improve the long-term planning ability of VINs. 
We improve traditional VINs by incorporating three key components: 
an aggregate gate, which establishes skip connections and facilitates long-term credit assignment; 
an exploration module, crafted to diversify information and gradient flow between neurons;
and a filter gate, designed to eliminate non-essential information.
Our experiments demonstrate that highway VINs enable long-term planning by training neural networks with hundreds of layers, surpassing the performance of several advanced methods. Future research will investigate the integration of multiple parallel VE modules with various types of embedded policies to improve performance. Additionally, future work will focus on scaling up to larger tasks.

\section*{Acknowledgement}

We sincerely thank the authors of the GPPN paper \cite{lee2018gated} for providing experimental details and Prof. Chao Huang for his valuable suggestions.
This work was supported by the European Research Council (ERC, Advanced Grant Number 742870) and the Swiss National Science Foundation (SNF, Grant Number 200021 192356).

\section*{Impact Statement}
This paper introduces research aimed at advancing the field of machine learning. While acknowledging numerous potential societal consequences, we believe that none need to be specifically emphasized here.

\bibliography{lib_my/bib,lib_my/bib_juergen}
\bibliographystyle{icml2024}

\newpage
\appendix
\onecolumn

\renewcommand{\thetheorem}{\thesection.\arabic{theorem}}
\setcounter{theorem}{0}

\section{Experimental Details}\label{sec_experiment_detail_app}
Our experimental settings follow those outlined in the paper on GPPN \cite{lee2018gated}.
For maze navigation tasks, the training, validation, and test datasets comprise $25K$, $5K$, and $5K$ mazes, respectively. 

All models are trained for $30$ epochs using the RMSprop optimizer with a learning rate of $0.001$ and a batch size of $32$. We also specify a kernel size of $5$ for convolutional operations in the planning module, as mentioned in \Cref{eq_Q_from_V}.
For the neural network that maps the observation to the latent MDP, we set the hidden dimension to $150$.

\section{Additional Experimental Results}\label{sec_experiment_result_app}
\subsection{Various Depths of Highway VIN}
\begin{figure*}[b]
    \centering
    \def\widthAJEJJDJ{0.24}
 \centerline{
  \subfloat[VIN]{
    \includegraphics[width=\widthAJEJJDJ\linewidth]{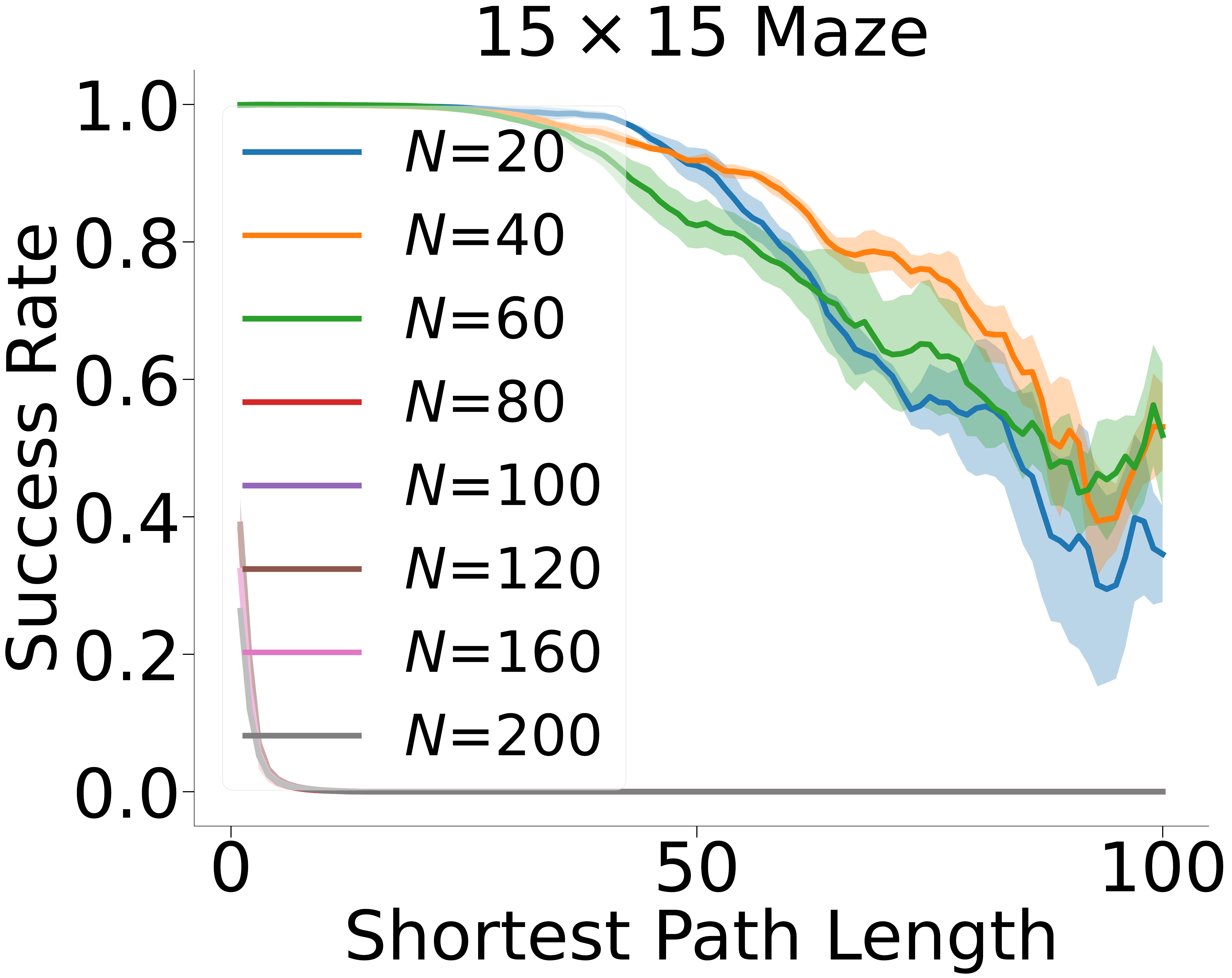}
    \includegraphics[width=\widthAJEJJDJ\linewidth]{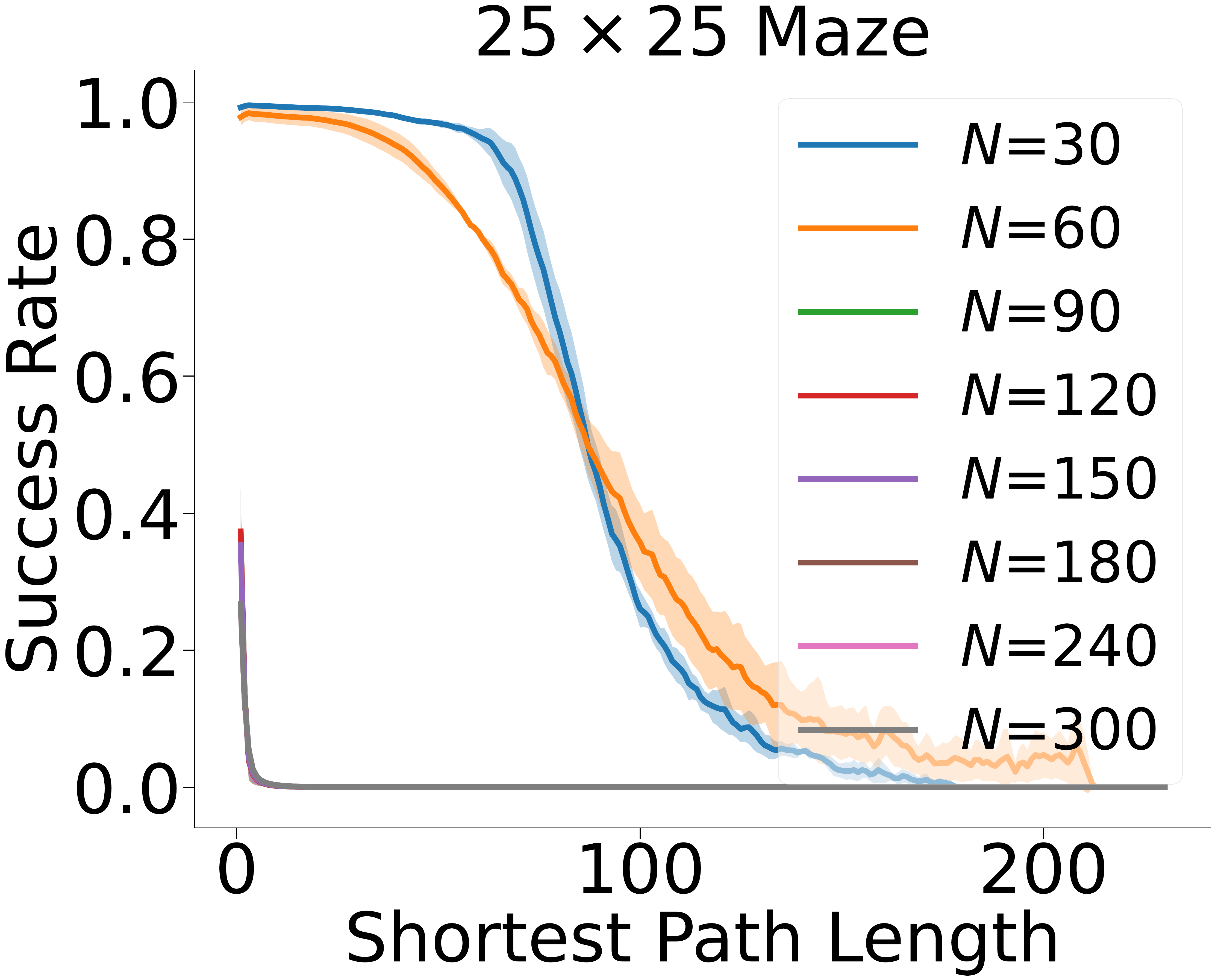}
 }
  \subfloat[Highway network]{
    \includegraphics[width=\widthAJEJJDJ\linewidth]{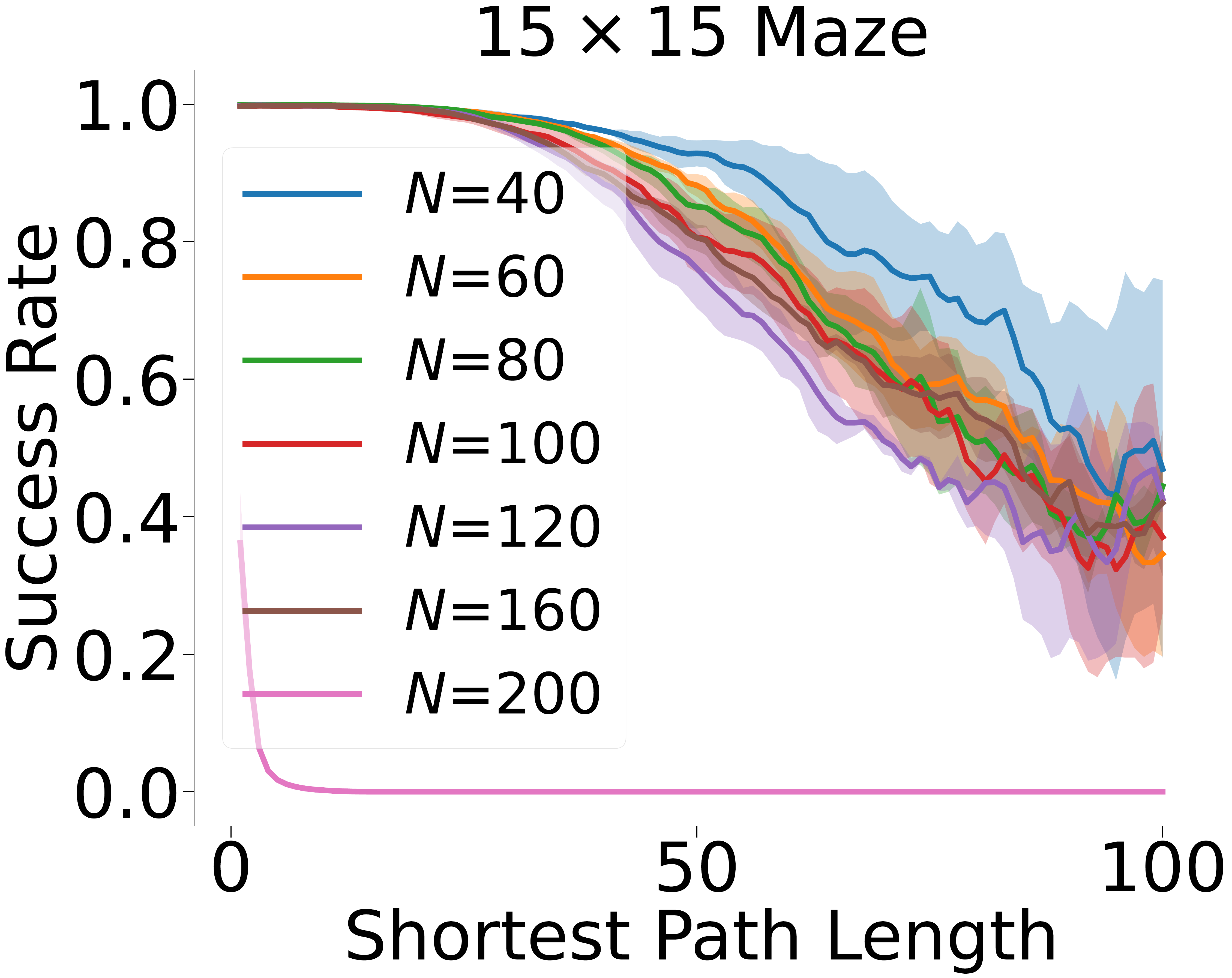}
    \includegraphics[width=\widthAJEJJDJ\linewidth]{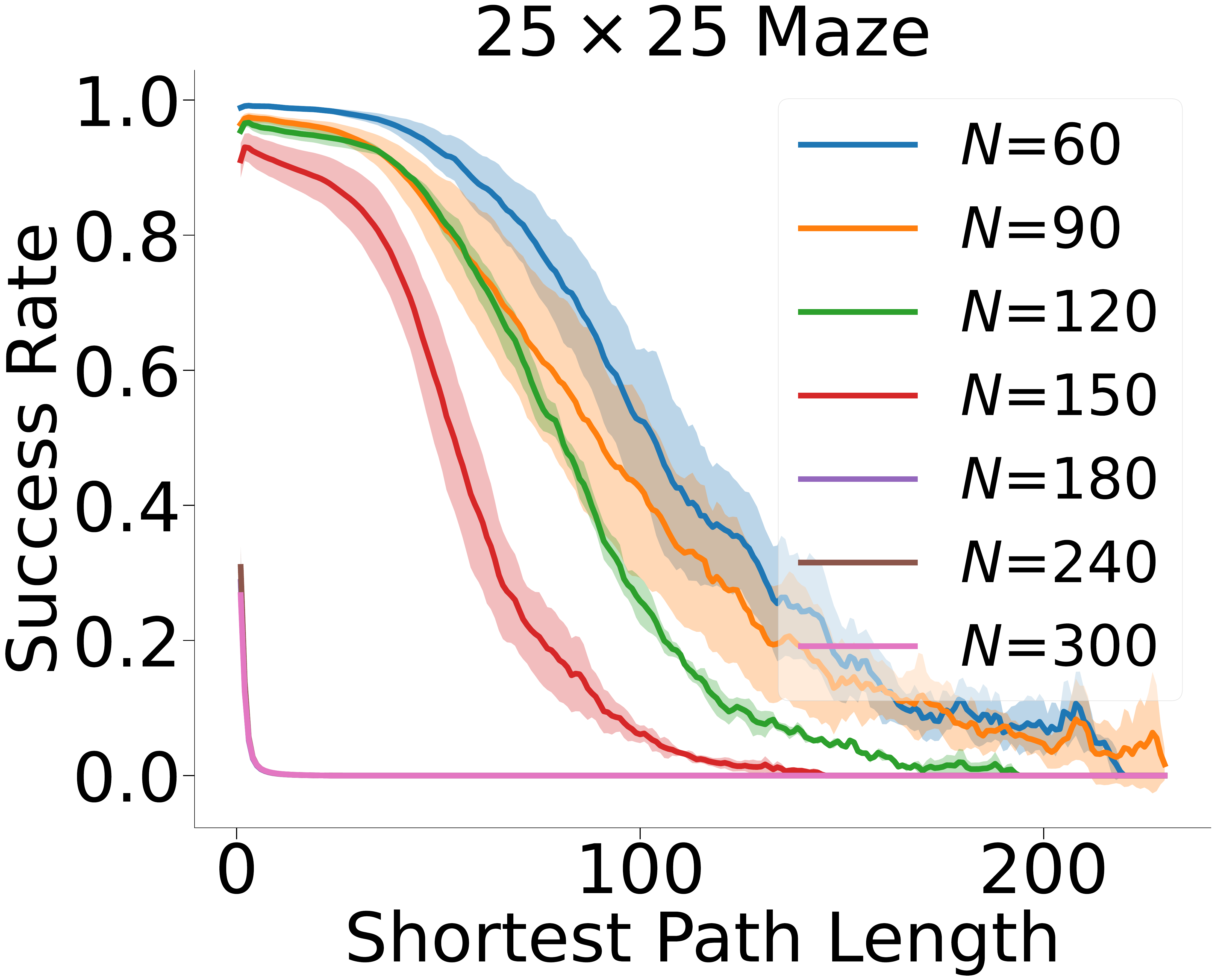}
 }
 }
 \centerline{
  \subfloat[GPPN]{
    \includegraphics[width=\widthAJEJJDJ\linewidth]{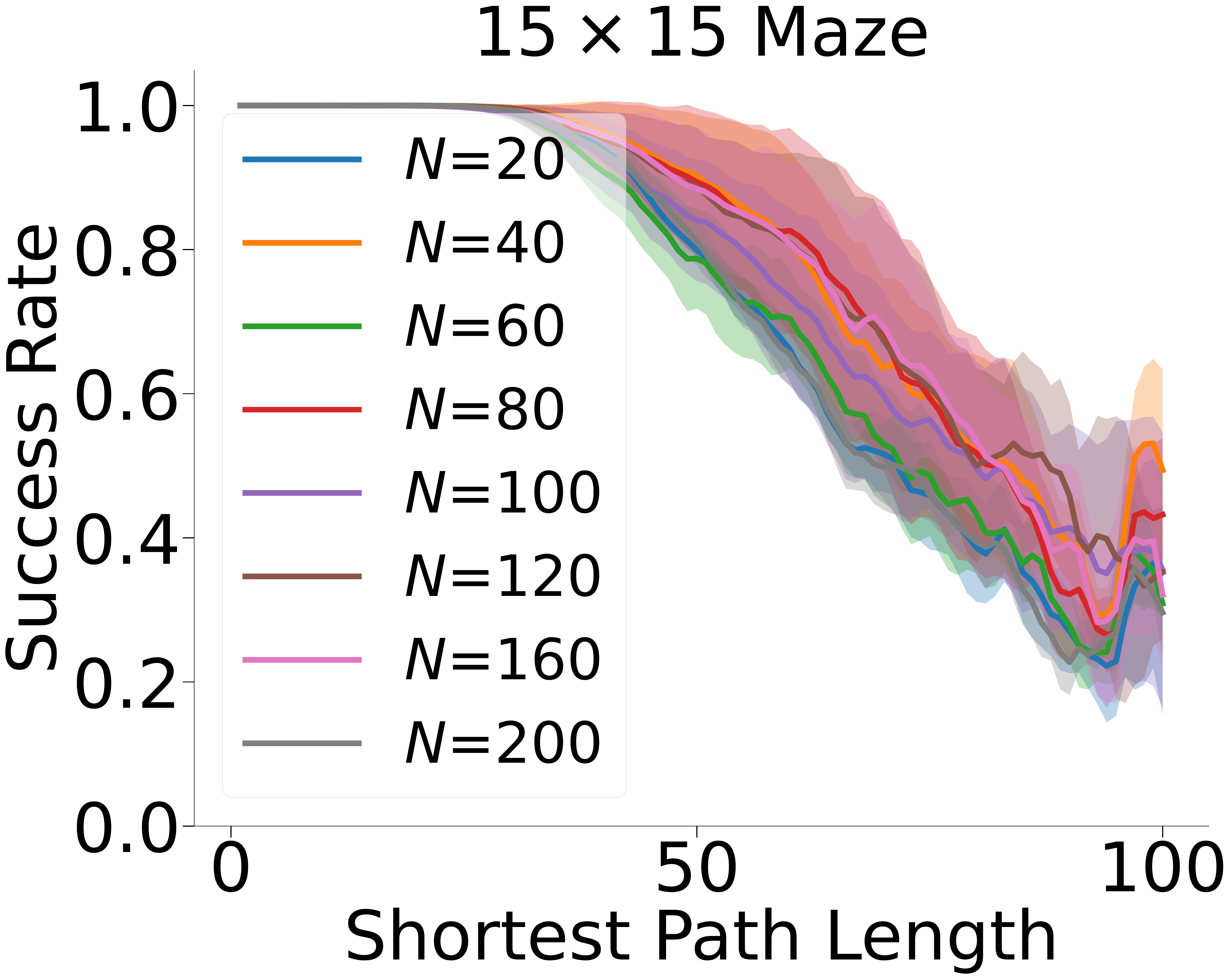}
    \includegraphics[width=\widthAJEJJDJ\linewidth]{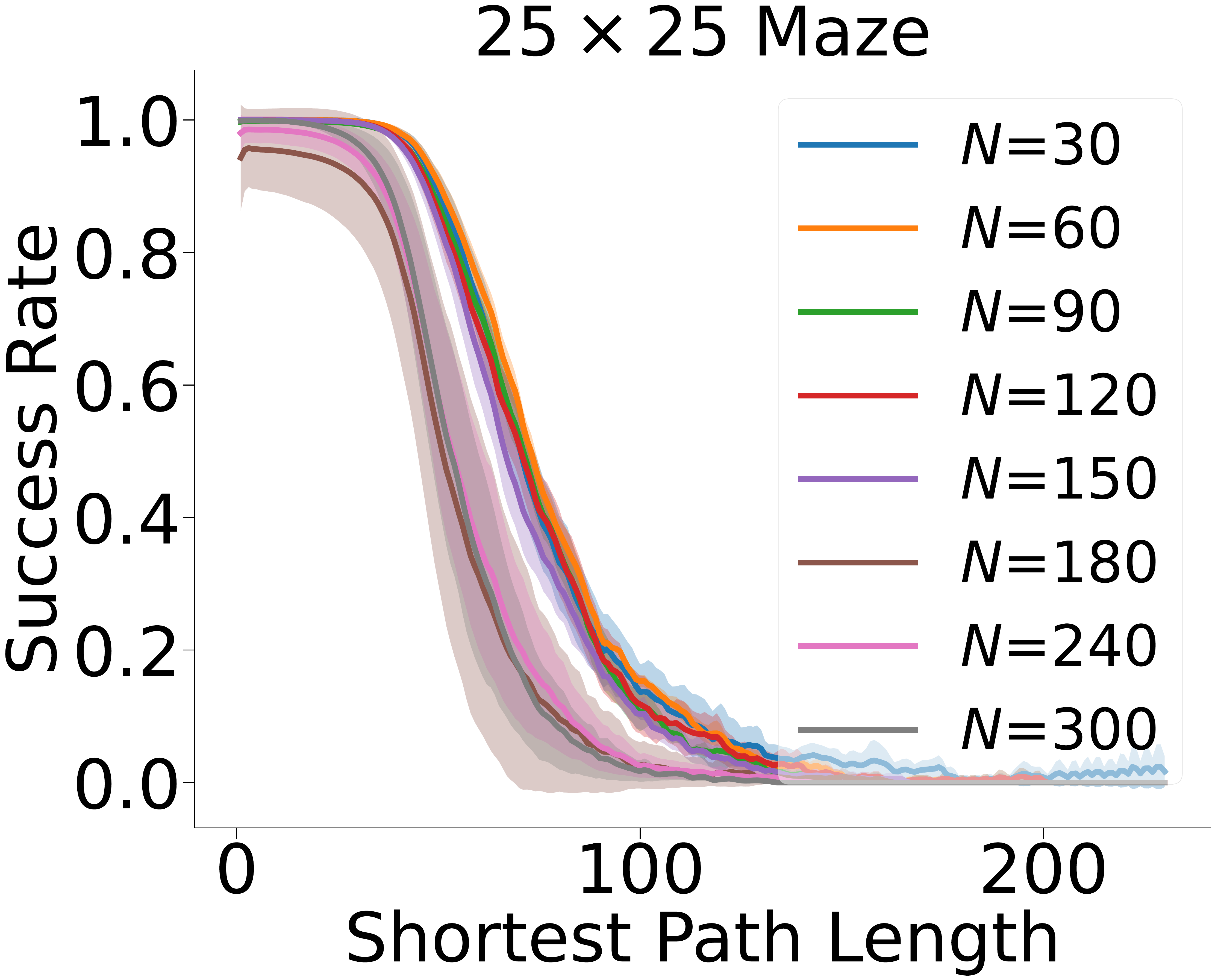}
 }
  \subfloat[Highway VIN (our)]{
    \includegraphics[width=\widthAJEJJDJ\linewidth]{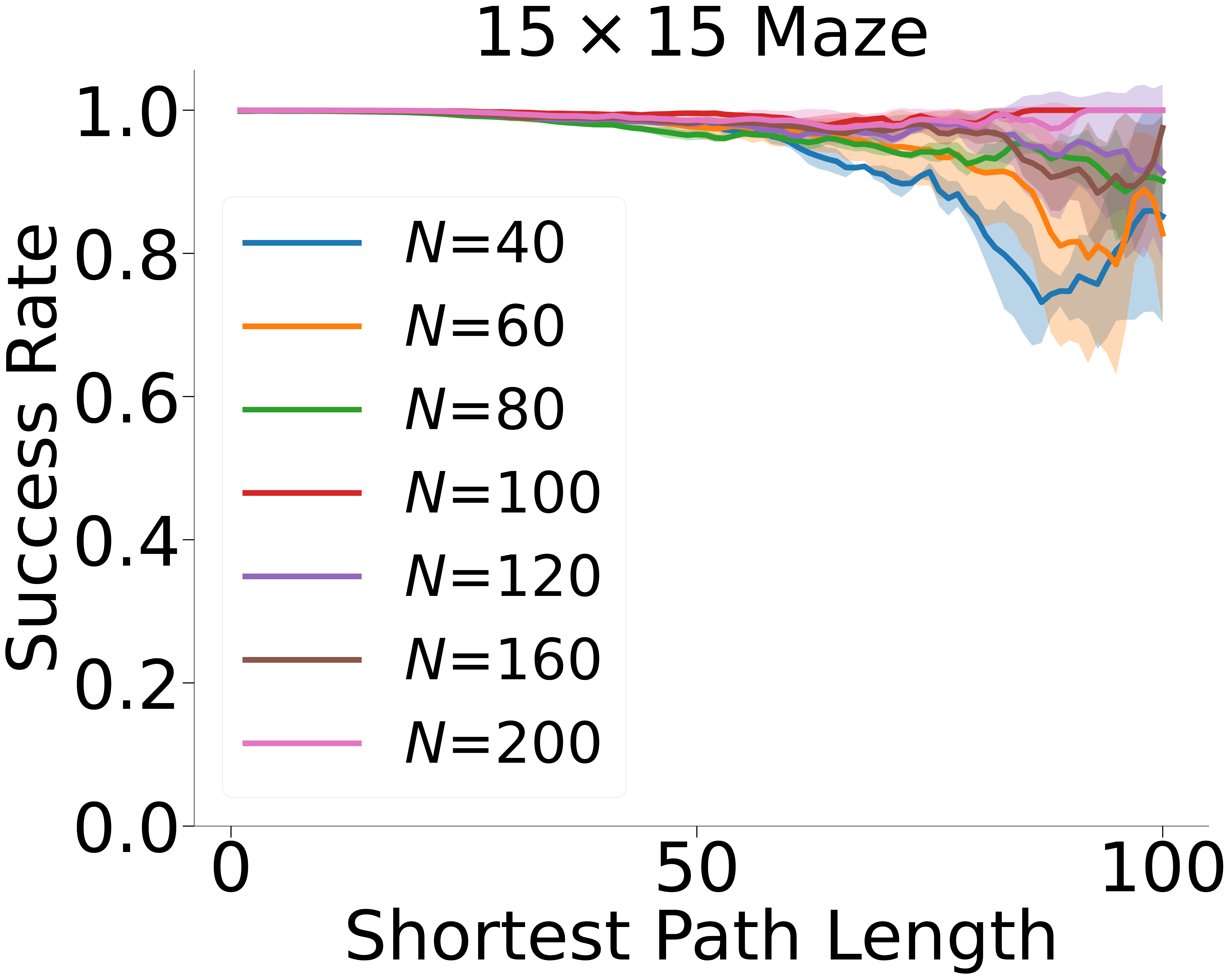}
    \includegraphics[width=\widthAJEJJDJ\linewidth]{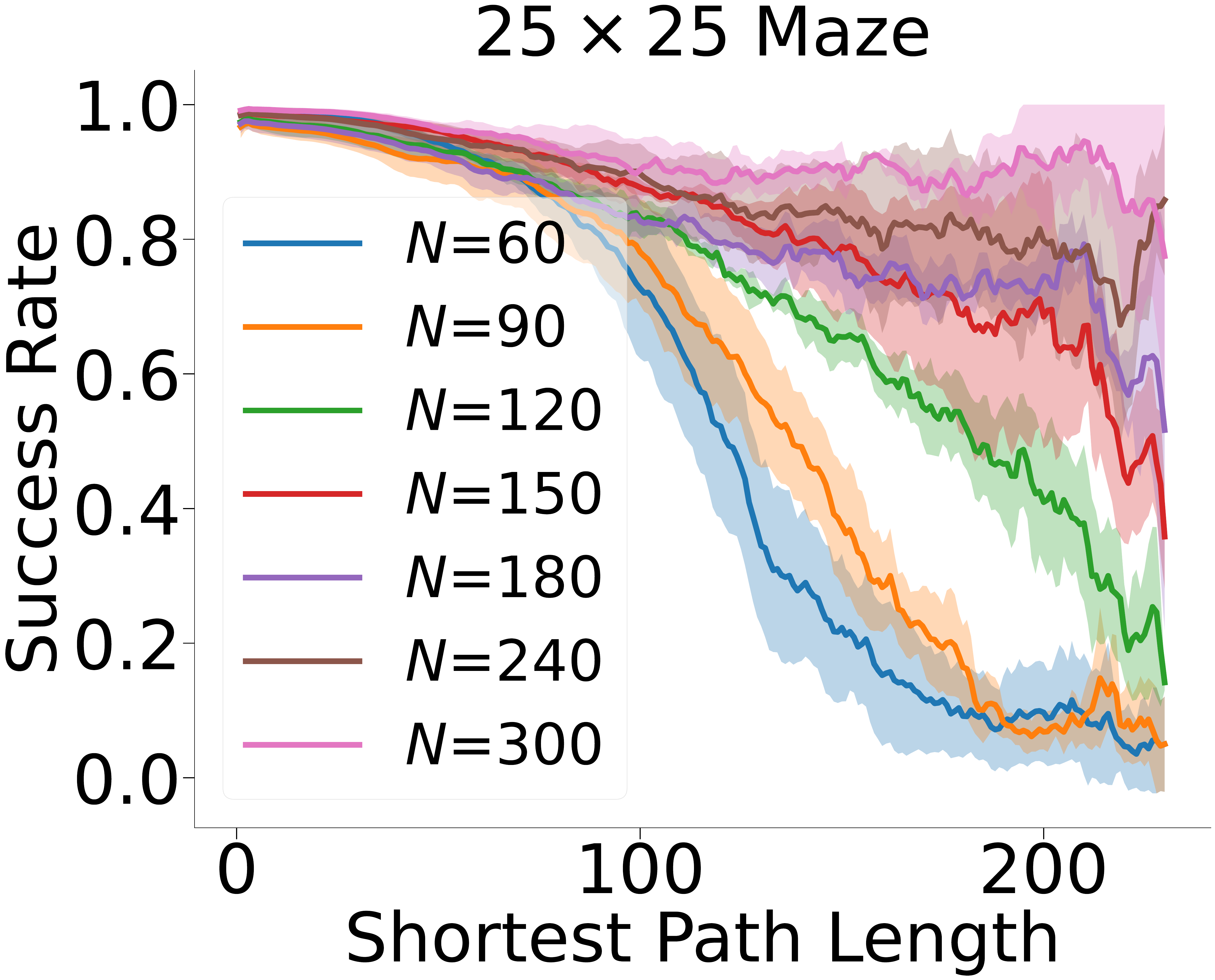}
 }
 }
    
    \caption{ 
    Success rates of each algorithm as a function of varying shortest path lengths on 2D maze navigation tasks.
    }
    \label{fig__success_rate__algorithm__all_depths}
\end{figure*}

\Cref{table_success_rate__all_depths} and \Cref{fig__success_rate__algorithm__all_depths} show the SRs of various algorithms across different depths.
For each algorithm, we also provide the rate at which it can plan the optimal path that yields the shortest path length, summarized in \Cref{table_optimal_rate}.

\subsection{Number of Parallel VE Modules}\label{sec_number_parrallel_VE_modules}

\begin{figure}[t]
    \centering
    \def\heightXAASD{0.22}
    \centerline{
        \subfloat[Depth $N=200$]{
            \includegraphics[height=\heightXAASD\linewidth]{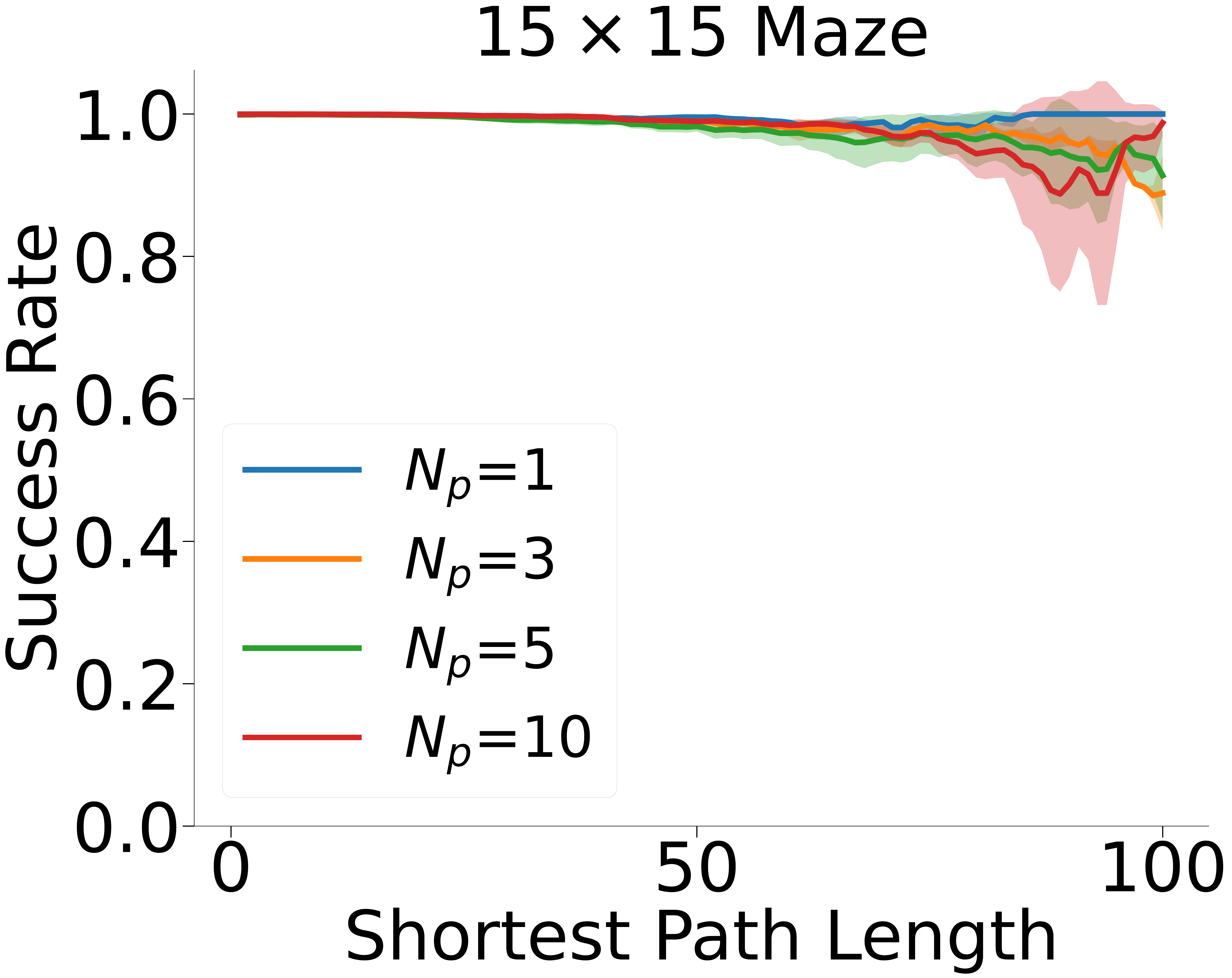} 
        }
        \subfloat[Depth $N=300$]{
            \includegraphics[height=\heightXAASD\linewidth]{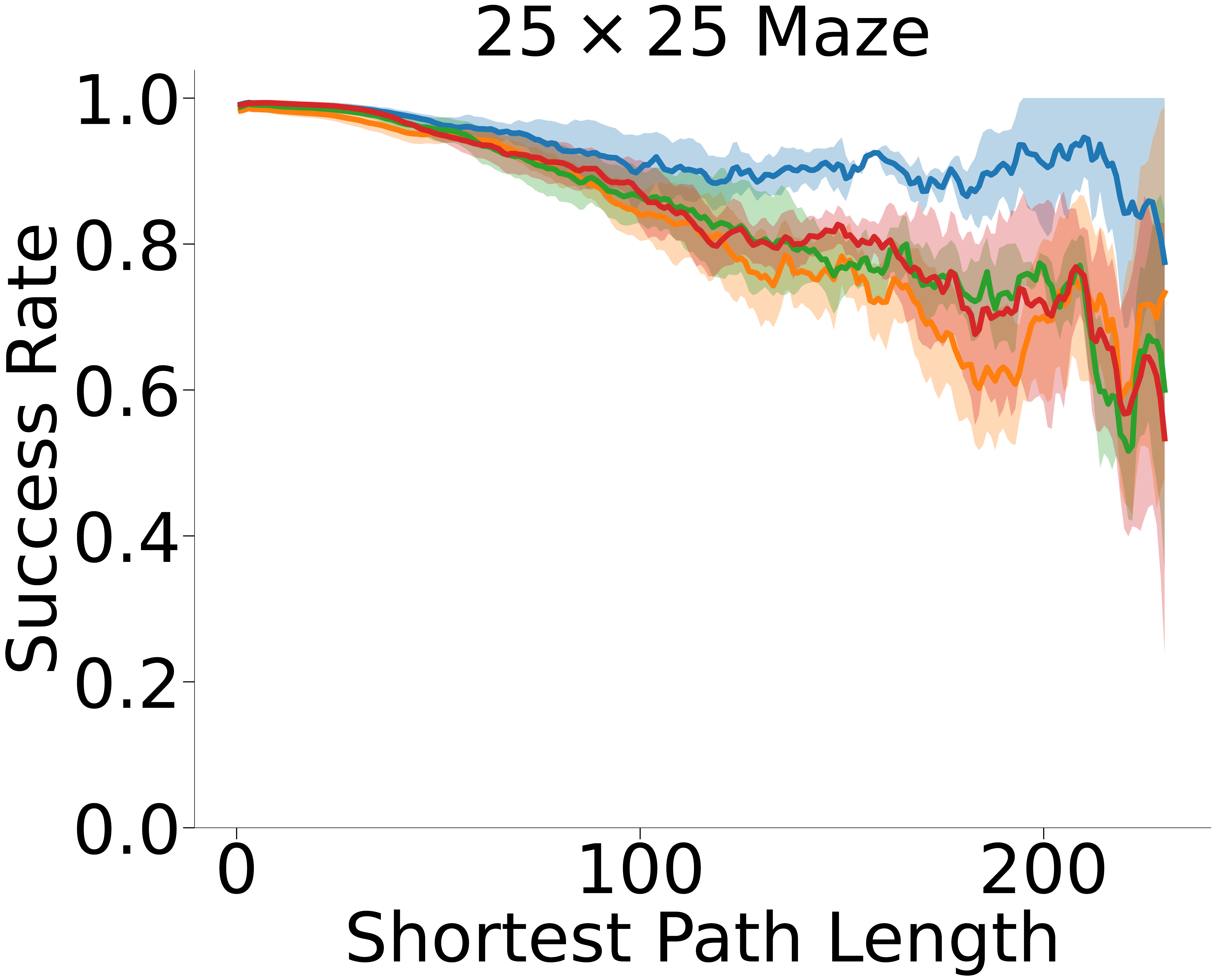}
        }
    }
    \centerline{
        \subfloat[Depth $N=100$]{
            \includegraphics[height=\heightXAASD\linewidth]{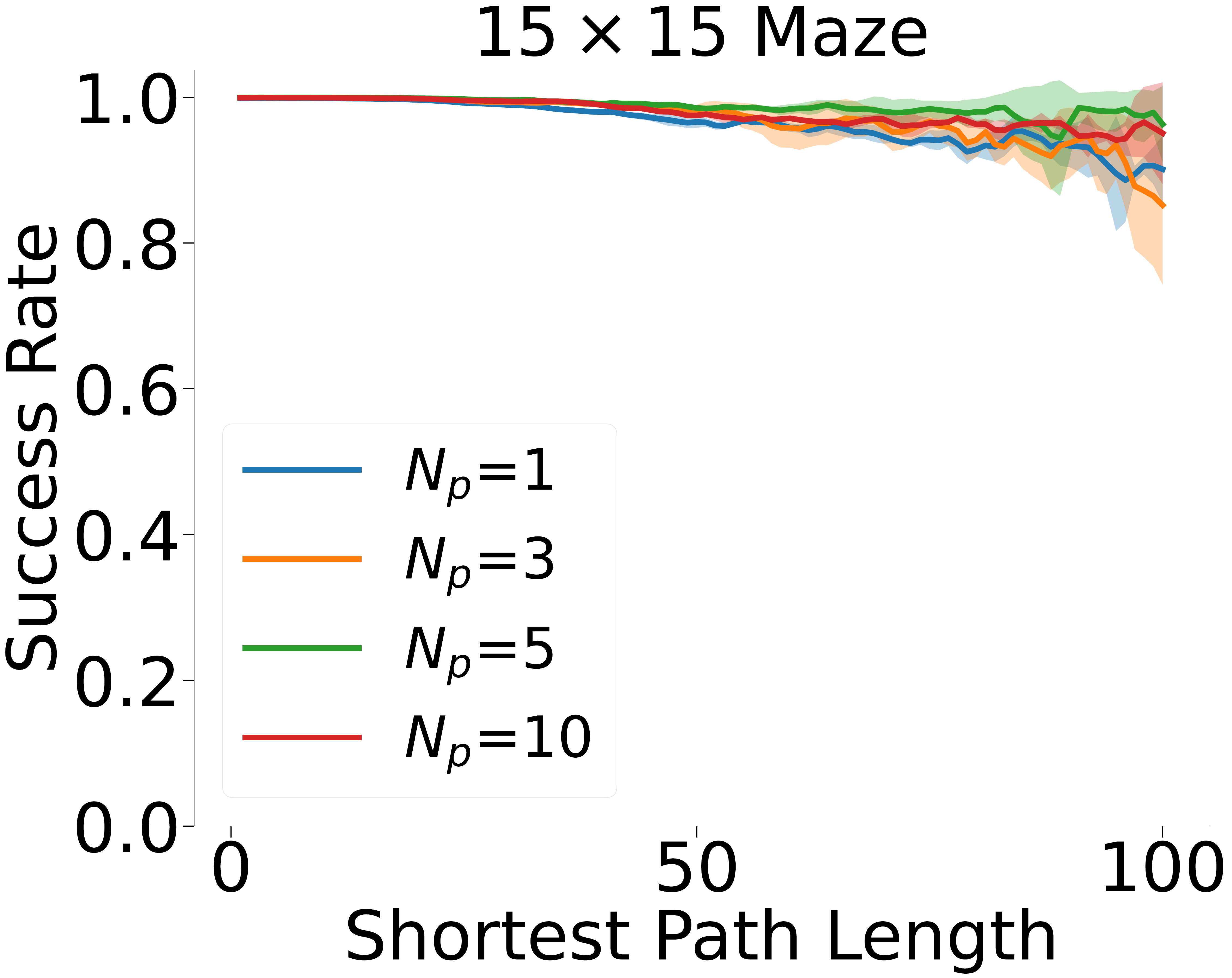} 
        }
        \subfloat[Depth $N=150$]{
            \includegraphics[height=\heightXAASD\linewidth]{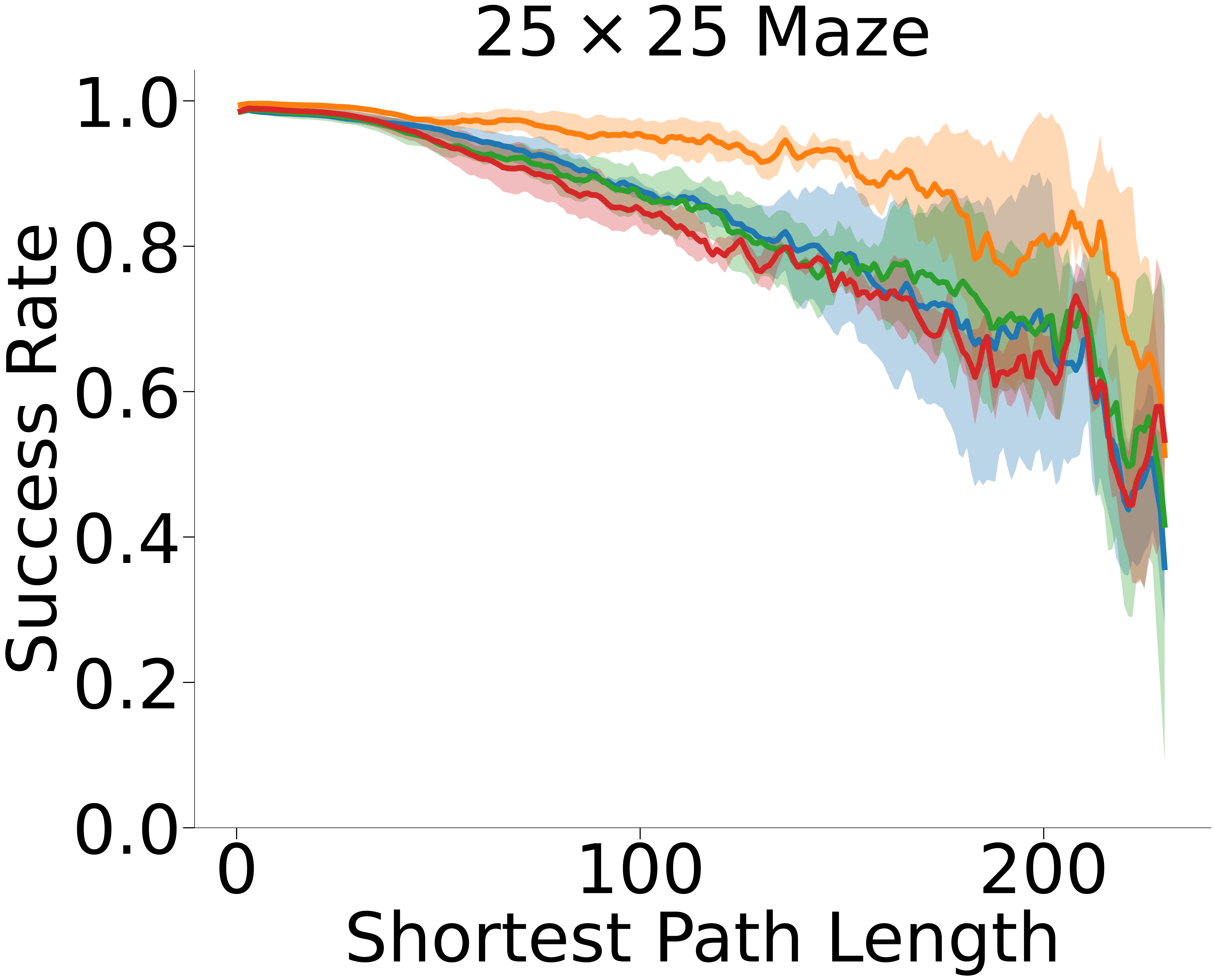}
        }
    }
    \caption{
    Success rates of highway VINs with varying numbers of parallel VE modules $\policyCount$ under varying depths $N$ of the network.
    }
    \label{fig_policy}
\end{figure}

We evaluate highway VIN with varying numbers of parallel VE modules $N_p$ under varying depths $N$.
As shown in \Cref{fig_policy}, under different depths $N$, the number of parallel VE modules has a different effect on the performance of highway VIN.
For example, under depth $N=300$, with fewer parallel VE modules, i.e.,  $N_p=1$, highway VIN performs the best. While under depth $N=100$, with more VE modules, $N_p=3$, highway VIN performs the best.
These results imply that the additional parallel VE modules may be detrimental to the performance of very deep networks.

\subsection{Examples of 2D Maze Navigation}
\begin{figure}[t]
    \centering
    \def\heightXAASD{0.33}
    \centerline{
        \subfloat[Highway VIN]{
            \includegraphics[width=\heightXAASD\linewidth]{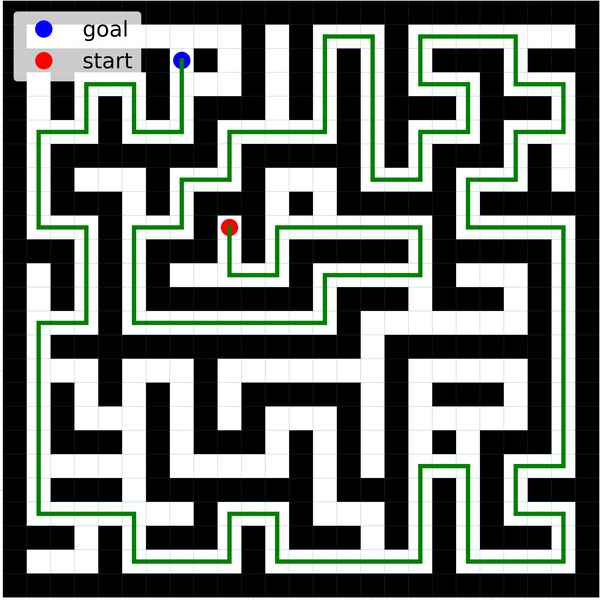} 
        }
        \subfloat[VIN]{
            \includegraphics[width=\heightXAASD\linewidth]{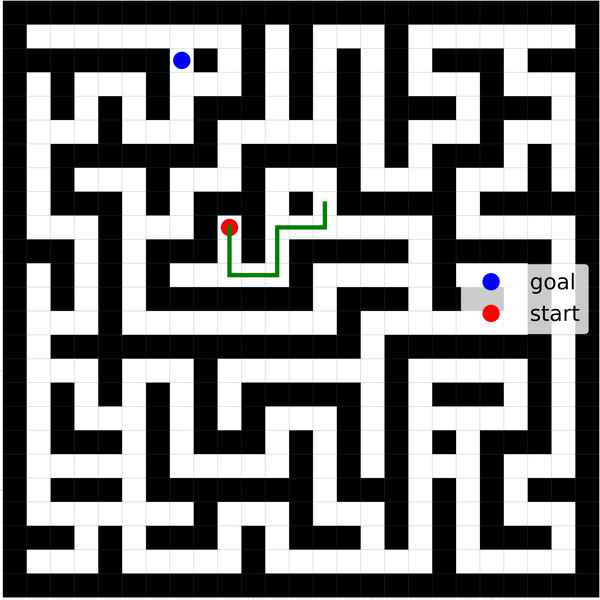}
        }
        \subfloat[GPPN]{
            \includegraphics[width=\heightXAASD\linewidth]{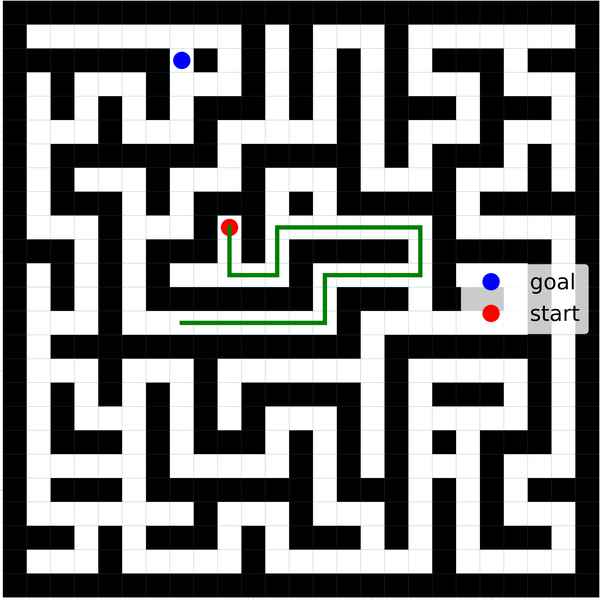}
        }
    }
    \centerline{
        \subfloat[Highway VIN]{
            \includegraphics[width=\heightXAASD\linewidth]{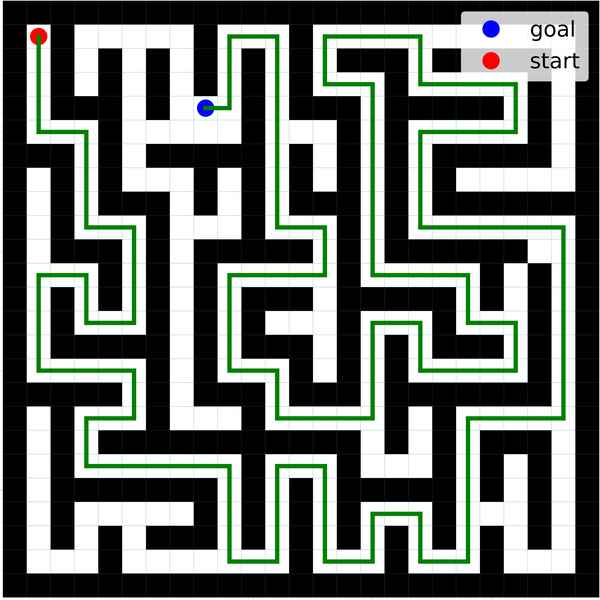} 
        }
        \subfloat[VIN]{
            \includegraphics[width=\heightXAASD\linewidth]{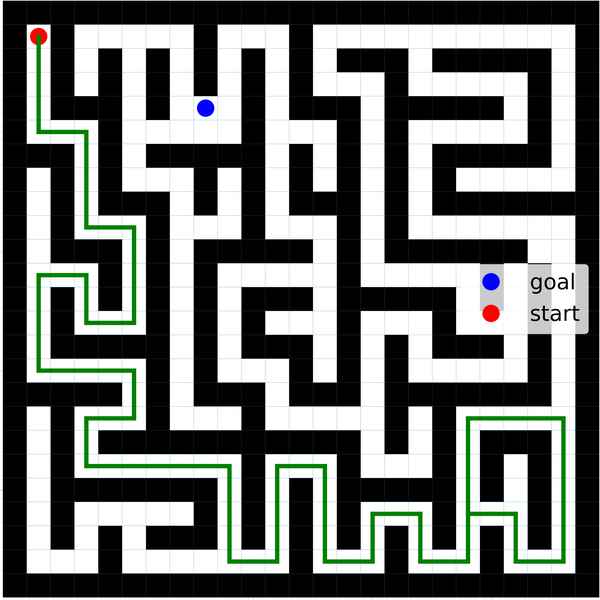}
        }
        \subfloat[GPPN]{
            \includegraphics[width=\heightXAASD\linewidth]{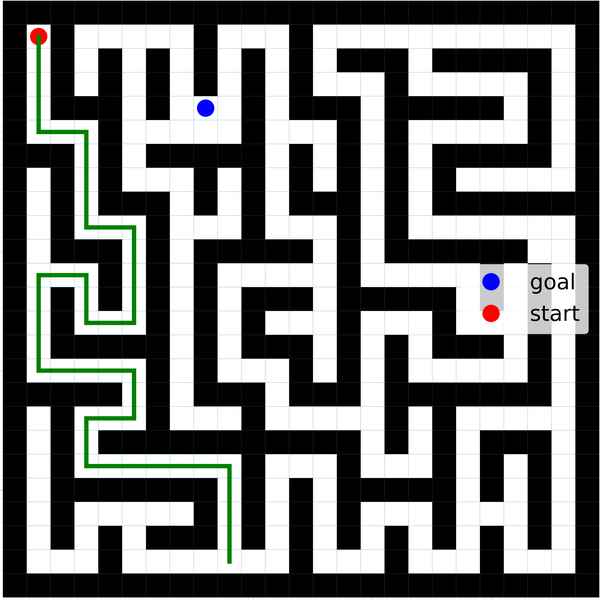}
        }
    }
    \caption{
    Examples of 2D maze navigation tasks where highway VIN succeeds, but other methods fail.
    }
    \label{fig_maze_examples}
\end{figure}

In \Cref{fig_maze_examples}, we show examples where highway VIN succeeds, but other methods fail.

\iftrue
\begin{table*}
\centering
\caption{
Success rates of each algorithm with various depths under 2D maze navigation tasks with different ranges of shortest path lengths.
\label{table_success_rate__all_depths}
}
\resizebox{\textwidth}{!}{
\begin{tabular}{c|l|c|c|c|l|c|c|c|}
\hline
Maze Size & \multicolumn{4}{c|}{$15\times 15$} & \multicolumn{4}{c|}{$25\times25$}\\ \hline
Shortest Path Length && $[1,30]$ & $[30,60]$ & $[60,100]$ && $[1,60]$ & $[60,130]$ & $[130,230]$ \\ \hline
\multirow{8}{*}{ \thead{VIN \\ \cite{tamar2016value}}} & $N=20$ & $99.83 \pm 0.11$ & $96.48 \pm 0.58$ & $63.03 \pm 3.20$ & $N=30$ & $98.84 \pm 0.16$ & $49.25 \pm 4.16$ & \phantom{9}$2.96 \pm 0.66$ \\
& $N=40$ & $99.79 \pm 0.10$ & $95.84 \pm 0.69$ & $76.16 \pm 1.87$ & $N=60$ & $96.47 \pm 1.33$ & $48.26 \pm 4.21$ & \phantom{9}$7.87 \pm 3.54$ \\
& $N=60$ & $99.83 \pm 0.03$ & $92.53 \pm 1.33$ & $66.18 \pm 6.91$ & $N=90$ & \phantom{9}$0.21 \pm 0.08$ & \phantom{9}$0.00 \pm 0.00$ & \phantom{9}$0.00 \pm 0.00$ \\
& $N=80$ & \phantom{9}$0.65 \pm 0.16$ & \phantom{9}$0.00 \pm 0.00$ & \phantom{9}$0.00 \pm 0.00$ & $N=120$ & \phantom{9}$0.21 \pm 0.08$ & \phantom{9}$0.00 \pm 0.00$ & \phantom{9}$0.00 \pm 0.00$ \\
& $N=100$ & \phantom{9}$0.80 \pm 0.03$ & \phantom{9}$0.00 \pm 0.00$ & \phantom{9}$0.00 \pm 0.00$ & $N=150$ & \phantom{9}$0.22 \pm 0.08$ & \phantom{9}$0.00 \pm 0.00$ & \phantom{9}$0.00 \pm 0.00$ \\
& $N=120$ & \phantom{9}$0.80 \pm 0.03$ & \phantom{9}$0.00 \pm 0.00$ & \phantom{9}$0.00 \pm 0.00$ & $N=180$ & \phantom{9}$0.24 \pm 0.00$ & \phantom{9}$0.00 \pm 0.00$ & \phantom{9}$0.00 \pm 0.00$ \\
& $N=160$ & \phantom{9}$0.64 \pm 0.12$ & \phantom{9}$0.00 \pm 0.00$ & \phantom{9}$0.00 \pm 0.00$ & $N=240$ & \phantom{9}$0.24 \pm 0.00$ & \phantom{9}$0.00 \pm 0.00$ & \phantom{9}$0.00 \pm 0.00$ \\
& $N=200$ & \phantom{9}$0.56 \pm 0.00$ & \phantom{9}$0.00 \pm 0.00$ & \phantom{9}$0.00 \pm 0.00$ & $N=300$ & \phantom{9}$0.24 \pm 0.00$ & \phantom{9}$0.00 \pm 0.00$ & \phantom{9}$0.00 \pm 0.00$ \\\hline
\multirow{8}{*}{\thead{GPPN \\ \cite{lee2018gated}}} & $N=20$ & $99.98 \pm 0.01$ & $92.68 \pm 1.07$ & $51.12 \pm 5.00$ & $N=30$ & $98.98 \pm 0.25$ & $25.98 \pm 5.78$ & \phantom{9}$2.76 \pm 1.68$ \\
& $N=40$ & ${\color{blue}\mathbf{99.99 \pm 0.01}}$ & $96.16 \pm 3.56$ & $65.17 \pm 12.4$ & $N=60$ & ${\color{blue}\mathbf{99.09 \pm 0.19}}$ & $28.87 \pm 1.47$ & \phantom{9}$1.32 \pm 0.55$ \\
& $N=60$ & $99.96 \pm 0.02$ & $91.47 \pm 3.50$ & $54.52 \pm 7.32$ & $N=90$ & $98.59 \pm 0.06$ & $25.35 \pm 2.66$ & \phantom{9}$0.86 \pm 0.59$ \\
& $N=80$ & $99.97 \pm 0.03$ & $95.44 \pm 4.48$ & $66.85 \pm 15.5$ & $N=120$ & $98.67 \pm 0.37$ & $25.60 \pm 4.87$ & \phantom{9}$1.35 \pm 0.98$ \\
& $N=100$ & $99.95 \pm 0.05$ & $93.34 \pm 4.16$ & $60.57 \pm 13.6$ & $N=150$ & $98.51 \pm 0.31$ & $21.62 \pm 3.50$ & \phantom{9}$0.73 \pm 0.68$ \\
& $N=120$ & ${\color{blue}\mathbf{99.99 \pm 0.01}}$ & $95.57 \pm 3.27$ & $66.99 \pm 15.2$ & $N=180$ & $90.49 \pm 8.62$ & \phantom{9}$7.40 \pm 8.56$ & \phantom{9}$0.41 \pm 0.58$ \\
& $N=160$ & $99.96 \pm 0.01$ & $95.51 \pm 3.13$ & $66.74 \pm 12.8$ & $N=240$ & $93.98 \pm 2.48$ & \phantom{9}$8.64 \pm 5.21$ & \phantom{9}$0.15 \pm 0.11$ \\
& $N=200$ & $99.98 \pm 0.01$ & $92.79 \pm 1.28$ & $50.88 \pm 3.59$ & $N=300$ & $95.38 \pm 2.01$ & \phantom{9}$6.29 \pm 4.35$ & \phantom{9}$0.02 \pm 0.03$ \\\hline
\multirow{7}{*}{\thead{Highway network \\ \cite{srivastava2015training}}} & $N=40$ & $99.65 \pm 0.17$ & $96.04 \pm 0.63$ & $75.86 \pm 10.0$ & $N=60$ & $97.93 \pm 0.56$ & $62.95 \pm 8.79$ & $17.46 \pm 5.45$ \\
& $N=60$ & $99.69 \pm 0.11$ & $94.31 \pm 0.55$ & $64.94 \pm 5.61$ & $N=90$ & $94.59 \pm 1.51$ & $49.91 \pm 11.8$ & $13.98 \pm 5.86$ \\
& $N=80$ & $99.70 \pm 0.05$ & $93.50 \pm 1.15$ & $62.22 \pm 5.87$ & $N=120$ & $93.65 \pm 0.81$ & $38.79 \pm 2.68$ & \phantom{9}$4.05 \pm 0.61$ \\
& $N=100$ & $99.36 \pm 0.32$ & $91.11 \pm 2.64$ & $60.32 \pm 8.87$ & $N=150$ & $85.42 \pm 4.20$ & $12.55 \pm 3.89$ & \phantom{9}$0.35 \pm 0.23$ \\
& $N=120$ & $99.51 \pm 0.17$ & $88.45 \pm 2.60$ & $51.88 \pm 4.24$ & $N=180$ & \phantom{9}$0.23 \pm 0.01$ & \phantom{9}$0.00 \pm 0.00$ & \phantom{9}$0.00 \pm 0.00$ \\
& $N=160$ & $99.50 \pm 0.05$ & $90.11 \pm 0.93$ & $60.57 \pm 3.33$ & $N=240$ & \phantom{9}$0.25 \pm 0.04$ & \phantom{9}$0.00 \pm 0.00$ & \phantom{9}$0.00 \pm 0.00$ \\
& $N=200$ & \phantom{9}$0.73 \pm 0.12$ & \phantom{9}$0.00 \pm 0.00$ & \phantom{9}$0.00 \pm 0.00$ & $N=300$ & \phantom{9}$0.24 \pm 0.00$ & \phantom{9}$0.00 \pm 0.00$ & \phantom{9}$0.00 \pm 0.00$ \\\hline
\multirow{7}{*}{ \thead{Highway VIN \\ (ours)}} & $N=40$ & $99.77 \pm 0.09$ & $98.83 \pm 0.25$ & $90.00 \pm 2.12$ & $N=60$ & $97.87 \pm 0.60$ & $77.02 \pm 6.30$ & $20.68 \pm 9.89$ \\
& $N=60$ & $99.83 \pm 0.10$ & $98.53 \pm 0.72$ & $94.35 \pm 4.15$ & $N=90$ & $95.31 \pm 1.69$ & $80.57 \pm 7.40$ & $34.72 \pm 6.27$ \\
& $N=80$ & $99.76 \pm 0.02$ & $98.03 \pm 0.02$ & $94.79 \pm 0.67$ & $N=120$ & $96.37 \pm 1.82$ & $84.81 \pm 2.12$ & $61.09 \pm 3.50$ \\
& $N=100$ & $99.93 \pm 0.03$ & ${\color{blue}\mathbf{99.52 \pm 0.12}}$ & ${\color{blue}\mathbf{98.61 \pm 0.66}}$ & $N=150$ & $97.77 \pm 0.48$ & $89.56 \pm 0.95$ & $75.42 \pm 10.1$ \\
& $N=120$ & $99.88 \pm 0.04$ & $98.62 \pm 0.35$ & $96.72 \pm 1.76$ & $N=180$ & $95.99 \pm 1.75$ & $85.18 \pm 2.28$ & $75.40 \pm 4.05$ \\
& $N=160$ & $99.86 \pm 0.04$ & $98.81 \pm 0.24$ & $96.76 \pm 1.02$ & $N=240$ & $97.64 \pm 1.49$ & $90.12 \pm 3.68$ & $82.40 \pm 8.95$ \\
& $N=200$ & $99.94 \pm 0.01$ & $99.13 \pm 0.12$ & $98.20 \pm 1.75$ & $N=300$ & $98.73 \pm 0.50$ & ${\color{blue}\mathbf{92.28 \pm 3.50}}$ & ${\color{blue}\mathbf{90.06 \pm 3.13}}$ \\\hline
\end{tabular}
}
\end{table*}

\begin{table*}[t]
\centering
\caption{
Optimality rates of each algorithm with various depths under 2D maze navigation tasks with different ranges of shortest path lengths. The optimality rate is defined by the ratio of tasks completed within the steps of the shortest path length to the total number of tasks.
\label{table_optimal_rate}
}
\resizebox{\textwidth}{!}{
\begin{tabular}{c|l|c|c|c|l|c|c|c|}
\hline
Maze Size & \multicolumn{4}{c|}{$15\times 15$} & \multicolumn{4}{c|}{$25\times25$}\\ \hline
Shortest Path Length && $[1,30]$ & $[30,60]$ & $[60,100]$ && $[1,60]$ & $[60,130]$ & $[130,230]$ \\ \hline
\multirow{8}{*}{ \thead{VIN \\ \cite{tamar2016value}}} & $N=20$ & $99.15 \pm 0.20$ & $90.50 \pm 0.59$ & $53.31 \pm 2.28$ & $N=30$ & $93.94 \pm 0.33$ & $38.32 \pm 3.64$ & \phantom{9}$2.25 \pm 0.35$ \\
& $N=40$ & $98.54 \pm 0.13$ & $86.71 \pm 0.56$ & $69.49 \pm 2.77$ & $N=60$ & $88.64 \pm 2.81$ & $33.74 \pm 3.27$ & \phantom{9}$6.16 \pm 2.38$ \\
& $N=60$ & $98.29 \pm 0.23$ & $81.58 \pm 2.57$ & $61.12 \pm 6.42$ & $N=90$ & \phantom{9}$0.20 \pm 0.09$ & \phantom{9}$0.00 \pm 0.00$ & \phantom{9}$0.00 \pm 0.00$ \\
& $N=80$ & \phantom{9}$0.61 \pm 0.16$ & \phantom{9}$0.00 \pm 0.00$ & \phantom{9}$0.00 \pm 0.00$ & $N=120$ & \phantom{9}$0.20 \pm 0.09$ & \phantom{9}$0.00 \pm 0.00$ & \phantom{9}$0.00 \pm 0.00$ \\
& $N=100$ & \phantom{9}$0.72 \pm 0.06$ & \phantom{9}$0.00 \pm 0.00$ & \phantom{9}$0.00 \pm 0.00$ & $N=150$ & \phantom{9}$0.21 \pm 0.09$ & \phantom{9}$0.00 \pm 0.00$ & \phantom{9}$0.00 \pm 0.00$ \\
& $N=120$ & \phantom{9}$0.72 \pm 0.06$ & \phantom{9}$0.00 \pm 0.00$ & \phantom{9}$0.00 \pm 0.00$ & $N=180$ & \phantom{9}$0.24 \pm 0.00$ & \phantom{9}$0.00 \pm 0.00$ & \phantom{9}$0.00 \pm 0.00$ \\
& $N=160$ & \phantom{9}$0.58 \pm 0.04$ & \phantom{9}$0.00 \pm 0.00$ & \phantom{9}$0.00 \pm 0.00$ & $N=240$ & \phantom{9}$0.24 \pm 0.00$ & \phantom{9}$0.00 \pm 0.00$ & \phantom{9}$0.00 \pm 0.00$ \\
& $N=200$ & \phantom{9}$0.56 \pm 0.00$ & \phantom{9}$0.00 \pm 0.00$ & \phantom{9}$0.00 \pm 0.00$ & $N=300$ & \phantom{9}$0.24 \pm 0.00$ & \phantom{9}$0.00 \pm 0.00$ & \phantom{9}$0.00 \pm 0.00$ \\\hline
\multirow{8}{*}{\thead{GPPN \\ \cite{lee2018gated}}} & $N=20$ & $99.35 \pm 0.12$ & $83.42 \pm 2.24$ & $46.97 \pm 5.81$ & $N=30$ & $96.33 \pm 0.33$ & $19.94 \pm 4.66$ & \phantom{9}$2.46 \pm 1.50$ \\
& $N=40$ & ${\color{blue}\mathbf{99.64 \pm 0.16}}$ & $90.47 \pm 6.65$ & $62.09 \pm 12.1$ & $N=60$ & ${\color{blue}\mathbf{96.53 \pm 0.65}}$ & $21.20 \pm 1.05$ & \phantom{9}$1.01 \pm 0.58$ \\
& $N=60$ & $99.36 \pm 0.18$ & $82.08 \pm 4.75$ & $49.27 \pm 7.18$ & $N=90$ & $94.78 \pm 0.21$ & $17.96 \pm 2.25$ & \phantom{9}$0.66 \pm 0.66$ \\
& $N=80$ & $99.41 \pm 0.24$ & $88.89 \pm 7.41$ & $61.21 \pm 13.3$ & $N=120$ & $95.44 \pm 0.38$ & $18.97 \pm 3.67$ & \phantom{9}$1.22 \pm 0.89$ \\
& $N=100$ & $99.27 \pm 0.27$ & $84.47 \pm 5.51$ & $55.82 \pm 12.6$ & $N=150$ & $95.05 \pm 0.66$ & $15.52 \pm 2.88$ & \phantom{9}$0.70 \pm 0.65$ \\
& $N=120$ & $99.48 \pm 0.12$ & $88.22 \pm 4.97$ & $64.14 \pm 14.6$ & $N=180$ & $82.70 \pm 11.6$ & \phantom{9}$5.40 \pm 6.49$ & \phantom{9}$0.40 \pm 0.56$ \\
& $N=160$ & $99.26 \pm 0.19$ & $85.70 \pm 3.63$ & $60.75 \pm 11.1$ & $N=240$ & $87.76 \pm 3.64$ & \phantom{9}$5.82 \pm 3.55$ & \phantom{9}$0.12 \pm 0.08$ \\
& $N=200$ & $99.38 \pm 0.08$ & $84.65 \pm 2.02$ & $47.10 \pm 3.90$ & $N=300$ & $88.50 \pm 4.61$ & \phantom{9}$4.10 \pm 2.93$ & \phantom{9}$0.02 \pm 0.03$ \\\hline
\multirow{7}{*}{\thead{Highway network \\ \cite{srivastava2015training}}} & $N=40$ & $98.98 \pm 0.12$ & $89.37 \pm 0.73$ & $71.27 \pm 9.95$ & $N=60$ & $92.57 \pm 1.68$ & $46.95 \pm 6.95$ & $13.72 \pm 4.40$ \\
& $N=60$ & $98.85 \pm 0.06$ & $85.92 \pm 0.31$ & $59.57 \pm 6.76$ & $N=90$ & $85.25 \pm 3.39$ & $35.28 \pm 9.05$ & $10.88 \pm 4.04$ \\
& $N=80$ & $98.62 \pm 0.19$ & $83.36 \pm 0.17$ & $56.04 \pm 4.09$ & $N=120$ & $83.27 \pm 0.19$ & $26.71 \pm 2.02$ & \phantom{9}$2.48 \pm 0.57$ \\
& $N=100$ & $97.88 \pm 0.24$ & $80.06 \pm 3.08$ & $55.21 \pm 9.75$ & $N=150$ & $71.57 \pm 5.70$ & \phantom{9}$7.47 \pm 2.58$ & \phantom{9}$0.10 \pm 0.10$ \\
& $N=120$ & $97.89 \pm 0.33$ & $78.04 \pm 2.05$ & $46.05 \pm 4.48$ & $N=180$ & \phantom{9}$0.23 \pm 0.01$ & \phantom{9}$0.00 \pm 0.00$ & \phantom{9}$0.00 \pm 0.00$ \\
& $N=160$ & $97.88 \pm 0.42$ & $78.57 \pm 0.95$ & $54.11 \pm 3.72$ & $N=240$ & \phantom{9}$0.25 \pm 0.04$ & \phantom{9}$0.00 \pm 0.00$ & \phantom{9}$0.00 \pm 0.00$ \\
& $N=200$ & \phantom{9}$0.65 \pm 0.09$ & \phantom{9}$0.00 \pm 0.00$ & \phantom{9}$0.00 \pm 0.00$ & $N=300$ & \phantom{9}$0.24 \pm 0.00$ & \phantom{9}$0.00 \pm 0.00$ & \phantom{9}$0.00 \pm 0.00$ \\\hline
\multirow{7}{*}{\thead{Highway VIN \\ (ours)}} & $N=40$ & $98.78 \pm 0.04$ & ${\color{blue}\mathbf{92.81 \pm 0.74}}$ & $85.49 \pm 2.83$ & $N=60$ & $93.47 \pm 0.67$ & $62.70 \pm 7.87$ & $17.15 \pm 8.31$ \\
& $N=60$ & $98.47 \pm 0.12$ & $91.46 \pm 1.55$ & $88.67 \pm 3.33$ & $N=90$ & $89.72 \pm 2.16$ & $63.87 \pm 8.26$ & $27.84 \pm 5.12$ \\
& $N=80$ & $98.62 \pm 0.23$ & $91.29 \pm 0.50$ & $90.99 \pm 0.41$ & $N=120$ & $90.41 \pm 1.76$ & $64.20 \pm 0.79$ & $49.63 \pm 2.57$ \\
& $N=100$ & $98.43 \pm 0.05$ & $90.67 \pm 0.51$ & ${\color{blue}\mathbf{94.64 \pm 1.61}}$ & $N=150$ & $92.00 \pm 0.58$ & $71.91 \pm 2.37$ & $64.70 \pm 9.88$ \\
& $N=120$ & $98.37 \pm 0.16$ & $90.24 \pm 1.31$ & $93.16 \pm 3.63$ & $N=180$ & $90.65 \pm 1.93$ & $66.42 \pm 1.83$ & $66.25 \pm 2.94$ \\
& $N=160$ & $98.30 \pm 0.11$ & $89.15 \pm 1.30$ & $92.00 \pm 0.44$ & $N=240$ & $91.32 \pm 2.24$ & $70.78 \pm 4.28$ & $71.09 \pm 8.04$ \\
& $N=200$ & $98.26 \pm 0.10$ & $89.33 \pm 0.92$ & $92.76 \pm 2.08$ & $N=300$ & $93.36 \pm 1.85$ & ${\color{blue}\mathbf{73.35 \pm 4.15}}$ & ${\color{blue}\mathbf{81.08 \pm 2.87}}$ \\\hline
\end{tabular}
}
\end{table*}
\fi
\end{document}